\documentclass[10pt,twocolumn,letterpaper]{article}

\usepackage[pagenumbers]{cvpr} % To force page numbers, e.g. for an arXiv version

\usepackage{times}
\usepackage{epsfig}
\usepackage{graphicx}
\usepackage{amsmath}
\usepackage{amssymb}

\usepackage[utf8]{inputenc}
\usepackage{comment}
\usepackage{times}
\usepackage{epsfig}
\usepackage{graphicx}
\usepackage{amsmath}
\usepackage{amssymb}
\usepackage{cuted}
\usepackage{capt-of}
\usepackage{adjustbox}
\usepackage{xcolor}
\usepackage{booktabs}

\newcommand{\lgh}[1]{\textcolor{blue}{#1}}

    \definecolor{marccolor}{rgb}{0,0.5,0.5}

    \definecolor{marccolorcom}{rgb}{0,0.8,0.8}

\definecolor{lghcom}{rgb}{0.3,0.3,0.8}

    \definecolor{mycolor}{rgb}{0.8,0.0,0.8}

\definecolor{cvprblue}{rgb}{0.21,0.49,0.74}
\usepackage[pagebackref,breaklinks,colorlinks,citecolor=cvprblue]{hyperref}

\def\paperID{7571} % *** Enter the Paper ID here
\def\confName{CVPR}
\def\confYear{2024}

\begin{document}

%\title{RealAvatar: Nonlinear Face Geometry Model and Complete Reflectance Map Synthesis from a Single Portrait Image}
%\title{SARD: Semi-Supervised Avatar Reconstruction with Differentiable Shading and Non-Linear Morphable Models}
\title{MoSAR: Monocular Semi-Supervised Model for Avatar Reconstruction using Differentiable Shading}

%%%%%%%%% AUTHORS - PLEASE UPDATE

\author
{\parbox{\textwidth}{\centering Abdallah Dib$^1$\thanks{Equal contribution}\;\;
                                Luiz Gustavo Hafemann$^{1*}$\;\;
                                Emeline Got$^1$\;\;
                                Trevor Anderson$^1$\;\;
                                Amin Fadaeinejad$^{1, 2 }$\;\;
                                Rafael M. O. Cruz$^3$\;\;
                                Marc-André Carbonneau$^1$
        }
        \\
        \\
{\parbox{\textwidth}{\centering  $^1$Ubisoft LaForge\;\;\; York University$^2$\;\;\;Ecole de Technologie Supérieure$^3$
       }
}
\vspace{-20px}
}

\maketitle
%\ificcvfinal\thispagestyle{empty}\fi

\begin{strip}\centering
%\vspace{-45px}
\includegraphics[width=\textwidth]{figures/teaser_small.pdf}
\captionof{figure}{
%Given a single image, our method achieves appealing 3D face reconstruction and estimates a dense detailed face geometry, spatially varying face reflectance (diffuse and specular albedos) and high frequency scene illumination.
Our method estimates detailed geometry and reflectance maps, yielding convincing rendering under new lighting conditions. 
\label{fig:teaser}}
\end{strip}
%%%%%%%%%%%%%%%%%%%%%%%%%%%%%%%%%%%%%%%%%%%%%%%%%%%%%%
% ABSTRACT
%%%%%%%%%%%%%%%%%%%%%%%%%%%%%%%%%%%%%%%%%%%%%%%%%%%%%%

%Reconstructing a fully-relightable avatar from a portrait image enables many applications in virtual reality, video games and VFX, but remains a challenging research problem. 

\begin{abstract}
Reconstructing an avatar from a portrait image has many applications in multimedia, but remains a challenging research problem. Extracting reflectance maps and geometry from one image is ill-posed: recovering geometry is a one-to-many mapping problem and reflectance and light are difficult to disentangle. Accurate geometry and reflectance can be captured under the controlled conditions of a light stage, but it is costly to acquire large datasets in this fashion. Moreover, training solely with this type of data leads to poor generalization with in-the-wild images. This motivates the introduction of MoSAR, a method for 3D avatar generation from monocular images. We propose a semi-supervised training scheme that improves generalization by learning from both light stage and in-the-wild datasets. This is achieved using a novel differentiable shading formulation. We show that our approach effectively disentangles the intrinsic face parameters, producing relightable avatars. As a result, MoSAR estimates a richer set of skin reflectance maps, and generates more realistic avatars than existing state-of-the-art methods. We also introduce a new dataset, named \emph{FFHQ-UV-Intrinsics}, the first public dataset providing intrinsic face attributes at scale (diffuse, specular, ambient occlusion and translucency maps) for a total of 10k subjects. The project website and the dataset are available on the following link: \url{https://ubisoft-laforge.github.io/character/mosar}

\end{abstract}

\section{Introduction}
\label{sec:intro}

Avatars are an important component of virtual worlds, being widely used in virtual reality, multimedia and video-games. Realistic personalized avatars, often called digital doubles, serve as bridge between the physical world and digital realities. The light stage \cite{debevec2012light} has long been regarded as the primary solution to obtain high-quality realistic avatars, however their high cost restricts their availability to the general public. They also require physical presence of users for scanning, which is impractical for many applications.

In contrast, creating avatars from a single monocular image enables a wide range of applications \cite{tewari17MoFA, tewari2018HighFidelity, tewari2019fml, dib2021practical, dib2023s2f2, feng2021learning, EMOCA:CVPR:2021,MICA:ECCV2022, chandran2021rendering, lei2023hierarchical}. 
These methods are fast and robust to arbitrary capture conditions, however, they fall short in terms of quality when compared to high-level multimedia production standards.

Moreover, these approaches often produce avatars that cannot seamlessly be integrated in tools used by content creators and artists, which limits the ability to edit the captured avatars. For example, modern graphic engines render realistic faces by relying on advanced materials and texture maps that these methods do not support. Even state-of-the-art methods \cite{yamaguchi2018high, lattas2020avatarme, lattas2023fitme, lattas2021avatarme++, papantoniou2023relightify} are limited to estimating diffuse, specular and normal maps, while modern engines use a broader range of maps including ambient occlusion and translucency to simulate global illumination and sub-surface scattering.

% creators and artists\lgh{, which} limits their ability

In this paper, we propose a new method for generating 3D avatars from a single monocular image. It estimates rich reflectance maps including ambient occlusion and translucency, in addition to the conventional diffuse, specular and normal maps, all at a 4K resolution. 
%This rich map set and high resolution is more aligned with modern rendering engines than other existing methods.  %LGH I think we already made this point above.
Our method can separate light contributions from the ambient occlusion and translucency from the diffuse maps because it relies on high-quality light stage data. It generalizes well to uncontrolled capture conditions because it also trains on a large quantity of in-the-wild monocular images using self-supervision. This is made possible by our new differentiable shading formulation. This paper makes the following main contributions:

\begin{itemize}
  
    \item We propose a semi-supervised training scheme leveraging high-quality light stage data, and large quantities of in-the-wild images. This allows for the estimation of a rich set of reflectance maps, as well as accurate geometry.
    \item We establish non-linear morphable models as a promising avenue to recover geometry from a single image by ranking in 2nd place on the REALY benchmark \cite{REALY}, so far largely dominated by linear models.
    \item We introduce a new differentiable shading formulation that incorporates, for the first time, the ambient occlusion and translucency maps with the spherical harmonics lighting model. 
    %\item We show that our method faithfully separates intrinsic face parameters from in-the-wild images. %Maybe say something like "a very challenging problem"
    
\end{itemize}

These contributions culminate in a more accurate and detailed head model than other state-of-the-art monocular avatar reconstruction methods. Figure \ref{fig:teaser} shows results from our method. In addition to our method's contribution, we introduce a new dataset, named \emph{FFHQ-UV-Intrinsics}, the first public dataset providing intrisic face attributes at scale (diffuse, specular, ambient occlusion and translucency maps) for a total of 10k subjects. We built it by applying our method to the dataset \emph{FFHQ-UV}\cite{Bai_2023_CVPR}.

\section{Related works}
\label{sec:sota}

\begin{figure*}
    \centering
    \includegraphics[width=\textwidth]{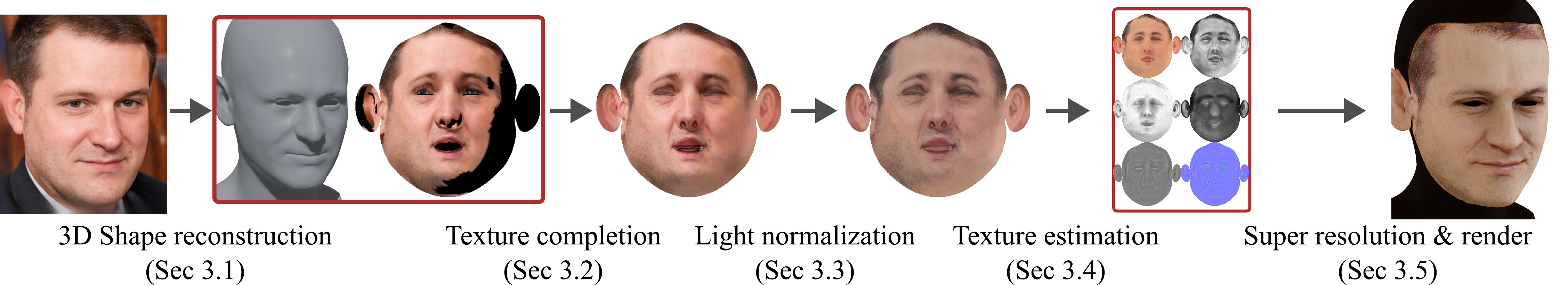}
    % \caption{Overview of the proposed pipeline. First, the \emph{3D shape reconstruction} module (section \ref{ssec:method_shape}) estimates the geometry and camera position, which is used to project the input image onto an UV space. Then, the \emph{texture completion} network (section \ref{ssec:method_tc}) estimates the missing parts of the texture. Next, light is normalized across the entire face (section \ref{ssec:method_ln}) and the \emph{face attributes} network (section \ref{ssec:method_td}) recovers various texture maps (i.e diffuse, specular, displacement, ambient occlusion and translucency). Finally, a \emph{super-resolution} network upscales the maps.}
    \caption{Overview of the proposed pipeline. In a first step, the 3D geometry of the face is estimated, and the image is projected onto the UV space. This texture is then used to estimate reflectance and displacement maps in 4K in a series of steps describe in Section \ref{sec:method}.}
    \label{fig:overview}
\end{figure*}
%\paragraph{3D shape reconstruction from images}
%\paragraph{3D face reconstruction}
\paragraph{3D Geometry reconstruction}
Face geometry estimation from images has been widely investigated in the literature \cite{zollhoefer2018facestar}. Accurate geometry estimation can be obtained in controlled capture settings with multiple cameras (e.g. in a light stage \cite{debevec2012light}), but this is costly and labor intensive, which makes it impractical for many use cases.
This motivated research on geometry recovery from a single image, which enables a wider variety of applications.

Monocular face reconstruction is commonly treated in a self-supervised way \cite{zollhoefer2018facestar}, by modeling a parametric scene with geometry, light, reflectance and camera parameters. These are used in conjunction with a differentiable renderer \cite{kato2020differentiable}, enabling learning from images only. There are three main types of approaches for the parametric geometry model: (i) using linear morphable models (3DMM), (ii) learning unconstrained geometry, and (iii) learning non-linear 3DMMs.

%\cite{blanz1999morphable,li2017learning,Egger20Years}  % check if we can cite these elsewhere
%Methods based on linear 3DMMs \cite{tewari17MoFA, feng2021learning, zhu2017face, deng2019accurate, danvevcek2022emoca, sanyal2019learning, dib2021towards} are restricted by the statistical prior space of 3DMM, and fall short of term of quality requirements expected in professional VFX and gaming pipeline. To go beyond the linear space of 3DMM, methods such as \cite{sela2017unrestricted,richardson2017learning,zeng2019df2net} estimate unrestricted geometry. However, the quality of the reconstructed mesh is limited and requires an additional registration process when used with a specific a template topology.
%\ma{The expressivity of methods based on linear 3DMMs \cite{tewari17MoFA, feng2021learning, zhu2017face, deng2019accurate, danvevcek2022emoca, sanyal2019learning, dib2021towards} is limited by the statistical prior space which often leads to inaccurate estimation. To go beyond the linear space of 3DMM, methods such as \cite{sela2017unrestricted,richardson2017learning,zeng2019df2net} estimate unrestricted geometry. However, so far, the quality of the reconstructed mesh is limited and requires an additional registration process when used with a specific a template topology.}

%Other methods focus on capturing detailed geometry \cite{chai2023hiface, chen2019photo, chen2020self, feng2021learning, lei2023hierarchical} and produces detailed geometry but neglect detailed albedo estimation which make them incomplete for production needs.

Methods based on linear 3DMMs \cite{tewari17MoFA, feng2021learning, zhu2017face, deng2019accurate, danvevcek2022emoca, sanyal2019learning, dib2021towards, chai2023hiface, lei2023hierarchical} are bound by the statistical prior space of the 3DMM,  which restricts their expressiveness. To go beyond this prior, methods such as \cite{sela2017unrestricted,richardson2017learning,zeng2019df2net} estimate unrestricted geometry.
These methods output dense meshes that require a registration process to a standard topology in order to be used in most applications. 
%However, the quality of the reconstructed mesh is limited \cite{dib2023s2f2} and requires an additional registration process when used with a specific a template topology.

Other methods such as \cite{tran2018nonlinear, tran2019towards} learn a non-linear morphable model. These methods train a decoder that estimates the final geometry under the weak supervision of another decoder that predicts 3DMM coefficients. The face geometry produced by these methods often contain artifacts as noted in \cite{dib2021towards}. 
Graph neural networks (GNN) have been used to refine the geometry obtained from 3DMMs \cite{lin2020towards, gao2020semi}, or to directly estimate geometry and reflectance \cite{lee2020uncertainty}. In \cite{aliari2023face}, a non-linear part-based GNN auto-encoder is trained for local shape editing. %While effective for artist-controlled editing, in this paper we show that it can also improve monocular face reconstruction.}

%Lin~\textit{et al.}\cite{lin2020towards} use graph neural networks to refine the estimated geometry obtained from 3DMM. Gao~\textit{et al.} \cite{gao2020semi} and Lee~\textit{et al.}\cite{lee2020uncertainty} use graph neural network to estimate geometry and reflectance. 
%Aliari~\textit{et al.} \cite{aliari2023face} train a non-linear part-based auto-encoder for local shape editing. While this method has been demonstrated to be effective in artist-controlled editing, we demonstrate in this paper that it can greatly improve monocular face reconstruction.

%Our work builds on top of \cite{lee2020uncertainty} and \cite{aliari2023face} by learning a part-based non-linear morphable model, not restricted nor regularized by linear 3DMMs, by leveraging GNN auto-encoders \cite{COMA:ECCV18}, trained in a semi-supervised setting. The proposed training scheme and loss functions greatly mitigate the problem of artifacts in the geometry, while still keeping it unbounded. %Additionally, our part-based GNN allows for efficient and local editing of the face which make it suitable for artists. %\ad{we need to say why we get better results than Tran et al. semi supervised learning? not regularized to 3DMM?}  

Our method is not restricted nor regularized by linear 3DMMs, and instead implements a non-linear morphable model using GNN auto-encoders \cite{COMA:ECCV18}, inspired by \cite{lee2020uncertainty} and \cite{aliari2023face}. Our proposed semi-supervised training scheme and loss functions mitigate artifact problems in the geometry, while still keeping it unbounded.

\paragraph{Skin reflectance}
To reconstruct a realistic and relightable avatar representation from an image, the intrinsic skin reflectance attributes must be disentangled. 
%Similarly, fine-grained geometry needs to be recovered. 
Recovering face attributes from single in-the-wild images is an ill-posed problem. Most existing methods \cite{tewari17MoFA, feng2021learning} use a Lambertian Bidirectional Reflectance Distribution Function (BRDF) \cite{cook1982reflectance} and a statistical basis, usually a PCA, for albedo estimation, which largely limits their expressivity.

% Recovering face attributes from in-the-wild image is an ill-posed problem. Most of existing methods

%Lee~\textit{et al.}\cite{lee2020uncertainty} learn an unbounded skin reflectance model from an input image. 
%However, the estimated texture map bakes light, geometry, and skin color (diffuse and specular) making it unsuitable for relighting or artist editing. 
%Our goal is to produce avatar with un-shaded albedos with disentangled attributes. 
%Dib et al. \cite{dib2023s2f2, dib2021towards, dib2021practical} model skin reflectance using a Cook-Torrance BRDF~\cite{cook1982reflectance,walter2007Microfacet}, recovering more details in the diffuse and specular components while being weakly regularized by a PCA. 
%These methods produce blurry textures, and do not capture fine-grained details of the skin reflectance. 
%Other methods (\cite{yamaguchi2018high,lattas2020avatarme,lattas2021avatarme++,lattas2023fitme,papantoniou2023relightify}) train a conditional GAN on light stage data to estimate diffuse and specular (Cook-Torrance), as well as normal maps for relightable avatars.

Some methods relax the constraint imposed by using a statistical basis. For instance in \cite{dib2023s2f2, dib2021towards, dib2021practical}, skin reflectance is modeled using a Microfacet BRDF \cite{torrance1967theory}, with diffuse and specular components recovered using only a weak PCA regularization. Nevertheless, this leads to blurry textures, and fails to capture fine-grained details of the skin. Lee~\textit{et al.}\cite{lee2020uncertainty} learn an unbounded skin reflectance model, 
however the estimated texture map bakes light, geometry and skin color (diffuse and specular) making it unsuitable for relighting. Finally, other methods \cite{yamaguchi2018high,lattas2020avatarme,lattas2021avatarme++,lattas2023fitme,papantoniou2023relightify} train a conditional GAN to estimate diffuse, specular, and normal maps for relightable avatars, but use only light stage data.

%In contrast to previous work, we estimate a richer skin reflectance, going beyond the Microfacet BRDF \cite{torrance1967theory} and including Ambient Occlusion and Translucency maps, all in 4K, enabling more realistic avatars, as well as being compatible with production-level artist workflows. Our model is trained in a semi-supervised way using a novel differentiable shading formulation, that enables improved results for in-the-wild images. This training scheme allows for disentangling diffuse from ambient occlusion resulting in more control over these parameters.

Our proposed method extends the Microfacet BRDF  and estimates a richer skin reflectance than existing methods. It produces Ambient Occlusion and Translucency maps, all in 4K, enabling more realistic avatars. These maps are also compatible with modern technical artist tools and rendering engines. Our semi-supervised training scheme, that uses a novel differentiable shading formulation, allows for disentangling diffuse from ambient occlusion, as well as specular and translucency maps, from in-the-wild images.

\section{Method}
\label{sec:method}

%Figure \ref{fig:overview} provides an overview of the  proposed method. Given an input image, the \emph{3D shape reconstruction} stage (section \ref{ssec:method_shape}) estimates geometry and camera, which is used to project the input image onto an UV space. The \emph{texture completion} network (section \ref{ssec:method_tc}) estimates the missing parts of the texture, and the \emph{light normalization} network (section \ref{ssec:method_ln}) ensures uniform light across the entire face region. The \emph{face attributes} network (section \ref{ssec:method_td}) recovers various texture maps encompassing diffuse, specular, displacement, ambient occlusion and translucency (sub-surface scattering) maps. Finally, a \emph{super-resolution} network upscales the maps to 4k.  This architecture is trained in a semi-supervised manner, using (i) lighstage data, that contains raw images, detailed geometry and 4K reflectance maps, and (ii) in-the-wild data, for which we only have face images.
%\commentlgh{It may be better to have this description at the caption of figure 2}
An overview of our method is shown in Figure \ref{fig:overview}. 
First, the \emph{3D shape reconstruction} module (Section \ref{ssec:method_shape}) estimates the geometry and camera position used to project the input image onto UV space. Then, the \emph{texture completion} network (Section \ref{ssec:method_tc}) estimates the missing parts of the texture. Next, light is normalized across the entire face (Section \ref{ssec:method_ln}) and the \emph{intrinsic texture map estimation} (Section \ref{ssec:method_td}) recovers various texture maps (i.e diffuse, specular, displacement, ambient occlusion and translucency). Finally, a \emph{super-resolution} network upscales the maps to 4K. 
We train this architecture in a semi-supervised manner, using (i) light stage data, that contains raw images, ground truth geometry and reflectance maps, and (ii) in-the-wild images, for which we do not have ground truth.

% geometry and camera position, used to project

%using a novel differentiable shading technique that incorporate AO and THK maps, compatible with the widely used spherical harmonics lighting model (section \ref{ssec:method_training}). 

\subsection{3D Shape Reconstruction}
\label{ssec:method_shape}

\begin{figure}
    \centering
    \includegraphics[width=\linewidth]{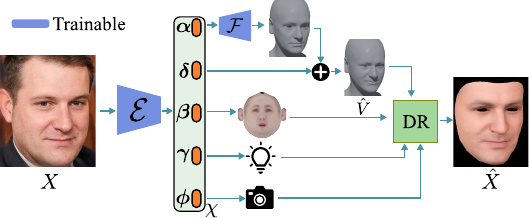}
    \caption{3D Face reconstruction pipeline: An encoder $\mathcal{E}$ estimates the scene parameters $\chi$, used by a Differential Renderer (DR) to obtain the estimated vertex colors, displayed here as an image.}
    \label{fig:3d}
\end{figure}

We represent the scene with a parametric model, defined by $\chi = \{ \boldsymbol{\alpha}, \boldsymbol{\delta}, \boldsymbol{\beta}, \boldsymbol{\gamma}, \boldsymbol{\phi} \}$, with neutral shape (identity) parameters $\boldsymbol{\alpha}$, expression coefficients $\boldsymbol{\delta}$, reflectance coefficients $\boldsymbol{\beta}$, light coefficients $\boldsymbol{\gamma}$ and camera parameters $\boldsymbol{\phi}$, as shown in Figure \ref{fig:3d}. %This decomposition is similar to previous work \cite{Garrido:2016, tewari17MoFA, dib2021practical}, with a novel training procedure for a non-linear representation for face identity.
This decomposition is similar to previous work \cite{Garrido:2016, tewari17MoFA, dib2021practical}, but we propose a novel procedure to learn a non-linear representation of identity.
%For the geometry reconstruction, we represent the scene with parameters $\chi = \{ \boldsymbol{\alpha}, \boldsymbol{\delta}, \boldsymbol{\beta}, \boldsymbol{\gamma}, \boldsymbol{\phi} \}$, with neutral shape coefficients $\boldsymbol{\alpha}$, expression coefficients $\boldsymbol{\delta}$, reflectance coefficients $\boldsymbol{\beta}$, light coefficients $\boldsymbol{\gamma}$ and camera parameters $\boldsymbol{\phi}$.

%For neutral shape, we use a part-based GNN model inspired by \cite{COMA:ECCV18, aliari2023face}. While this method has been demonstrated to be effective in artist-controlled editing, we demonstrate in this paper that it can greatly improve monocular face reconstruction. For the expression, we use linear blendshapes vectors. The vertex positions are computed as follows: $\mathsf{\hat{v}} = \mathcal{F}(\boldsymbol{\alpha}) + \sum_i{\delta_i \mathsf{E}_i}$,  where $\mathcal{F}(.)$ is the decoder from the GNN model and $\mathsf{E}_i$ is the $i$\textsuperscript{th} blendshape. In this work we use $\boldsymbol{\alpha} \in \mathbb{R}^{74}$ and $\boldsymbol{\delta} \in \mathbb{R}^{37}$.

%We obtain neutral shapes using a part-based Graph Neural Network (GNN) model inspired by \cite{COMA:ECCV18, aliari2023face} that generates detailed meshes.
We use a part-based GNN model inspired by \cite{COMA:ECCV18, aliari2023face} to generate a detailed neutral mesh.
%For modeling neutral meshes, we consider a part-based Graph Neural Network (GNN) model inspired by \cite{COMA:ECCV18, aliari2023face}. 
%While this method was proposed for artist-controlled geometry editing, we propose a semi-supervised training procedure that makes it effective for monocular face reconstruction, and shows a large improvement to linear 3DMMs.
While this method was proposed for artist-controlled geometry editing, our proposed semi-supervised training procedure makes it effective for monocular face reconstruction, improving modeling accuracy when compared to linear 3DMMs.

%The GNN model consists of part-based encoders $\mathcal{G}_i$, that encode meshes into latent representations $\alpha_i$ (where $i \in [0, \cdots, 7]$ correspond to different part of the face). 
The GNN model is composed of a collection of encoders $\mathcal{G}_i$ each responsible for translating the vertices of a specific part of the face into a latent representation $\boldsymbol{\alpha}_i$ where $i \in [0, \ldots, 7]$. These are concatenated in a single vector $\boldsymbol{\alpha}$, and fed to a decoder $\mathcal{F}(\boldsymbol{\alpha})$, that reconstructs the entire input mesh. We use the same architecture as \cite{aliari2023face}.

%The GNN model comprises a collection of encoders $\mathcal{G}_i$ each responsible for translating the vertices of a specific part of the face into a latent representations 

We represent facial expressions as a set of linear displacement blendshapes $\boldsymbol{E}$. Given parameters $\boldsymbol{\alpha}$ and $\boldsymbol{\delta}$, we obtain the geometry as $\hat{V} = \mathcal{F}(\boldsymbol{\alpha}) + \sum_i{\delta_i E_i}$, where $E_i$ is the $i$\textsuperscript{th} blendshape in the set.

For reflectance, we use a Lambertian BRDF model, with a PCA basis on per-vertex albedo colors.
%We represent the reflectance of an arbitrary face with vertex colors $\boldsymbol{\hat{c}} = \boldsymbol{a_r} + \sum_i{\beta_i A_i}$, where $\boldsymbol{a}_r$ is the average albedo and $\boldsymbol{A}$ are the eigenvectors of the PCA. 
%We use $\boldsymbol{\beta} \in \mathbb{R}^{200}$ coefficients and $\mathsf{\textbf{R}} \in \mathbb{R}^{200 \times 3N}$ for $N$ vertices in the mesh topology.
We model light using 3\textsuperscript{rd}-order spherical harmonics, and we use a pinhole camera model.
%: $\phi = \{ f,  \boldsymbol{t}, R \}$, with  focal length $f$, translation $\boldsymbol{t}$ and rotation $R$.
%
%

%
%

%
%
%\commentlgh{Not sure if we should cite MOFA here, it's not exactly the same terminology and can confuse the reader}
%For more details on the aforementioned parameters, we refer the reader to \cite{tewari17MoFA}.

%\commentma{This is extremly hard to follow: }

%Given an input image, a Resnet-based encoder $\mathcal{E}$ is used to encode the image into the latent representation $\chi$. 
%We use $\mathcal{F}(\alpha)$ to obtain the neutral mesh, and apply the expression basis according to the estimated coefficients $\delta$, obtaining vertex positions $\hat{v}$. 

\paragraph{Training}
We train two components at this stage: the GNN model producing neutral meshes and a scene-encoder $\mathcal{E}$ translating images into scene parameters $\chi$. We first pre-train the GNN model in a supervised way with the light stage data, following the work of \cite{aliari2023face}. Next, we train the scene encoder $\mathcal{E}$ and the GNN decoder $\mathcal{F}$ in a semi-supervised manner. We combine the in-the-wild data with light stage data, for which we have ground truth vertex positions used for supervision.

% We combine the in-the-wild data to light stage data

% camera $\boldsymbol{\phi}$
When training with in-the-wild data, we rely solely on self-supervision. The differentiable renderer uses the estimated geometry $\hat{V}$, reflectance coefficients $\boldsymbol{\beta}$, light coefficients $\boldsymbol{\gamma}$ and camera parameters $\boldsymbol{\phi}$ to compute the final color (irradiance) for each vertex, as in \cite{tewari17MoFA}. These are then compared to their corresponding pixels in the input image $X$. Additionally, we extract landmarks $l$ using an off-the-shelf detector \cite{bulat2017far}. During training, we minimize the following loss function: 
\begin{equation}
\label{eq:loss_geom}   
 \mathcal{L}_u =  L_\text{photo} + L_\text{landmark} + L_\text{reg}, 
\end{equation}
where the photo-consistency loss $L_\text{photo}$ is the $\ell_1$-norm between the predicted vertex colors and their associated pixel colors in the input image; $L_\text{landmark}$ is the $\ell_2$-norm of the distance between the landmarks $l$ and the perspective projection of their associated mesh vertices. $L_\text{reg}$ is the regularization term defined as: 
\begin{equation}
L_\text{reg} = L_\text{lap} + L_\text{light} + L_\text{exp} + L_\text{alb}.   
\end{equation}
$L_\text{lap}$ is the Laplacian smoothing operator from \cite{groueix20183d} applied to the estimated mesh $\hat{V}$, which minimizes the mean surface curvature.
$L_\text{exp}$ is the $\ell_1$-norm of the expression coefficient vector acting as a regularization term. $L_\text{light}$ encourages monochromatic light (similar to \cite{dib2021practical}), while $L_\text{alb}$ regularizes against implausible face deformations using prior albedo statistics as defined in \cite{tewari17MoFA}. 
We note that the term $L_\text{lap}$ if required since $\mathcal{F}$ is not restricted to a parametric model, and thus requires regularization to avoid producing implausible geometry.

% and thus require regularization to avoid producing unplausible geometry.

%This is because our non-linear geometric network is not restricted to a prior or parametric model (such as PCA). Consequently, the model may produces odd geometry if not regularized carefully with the Laplacian term.

When training with light stage data, we leverage the ground-truth mesh $v_i$ associated with the image $X_i$. In that case we complement $\mathcal{L}_u$ with $L_\text{supervised}$ and $L_\text{nrm}$ which are respectively the $\ell_2$-distance between the vertices of the estimated and GT mesh, and the cosine distance between their normal vectors:
\begin{equation}
\label{eq:loss_geom_sup}   
 \mathcal{L}_s =  \mathcal{L}_u + L_\text{supervised} + L_\text{nrm}.
\end{equation}

We omitted weighting coefficients ($\lambda$) of the loss term for better readability. They are specified in the supplementary material.

Finally, at inference time, we further refine the scene parameters for the input image. Similarly to \cite{dib2021practical}, we directly optimize the parameters $\chi$ to minimize Equation \ref{eq:loss_geom}, using the Adam optimizer \cite{kingma2014adam}, with the values given by the scene encoder $\chi = \mathcal{E}(X)$ as initialization.

\subsection{Texture completion}
\label{ssec:method_tc}

%\ad{add some references here for other existing works that perform texture completion}
%The estimated geometry and camera information are used to project the input image onto UV space. The resultant texture contains holes that corresponds to occluded or non visible areas of the face, as can be seen in Figure \ref{fig:overview}. The next step in our pipeline is to estimate the complete texture, similar to \cite{yamaguchi2018high, deng2018uv}. 

We project the image onto the UV space using the estimated geometry and camera information. This produces an incomplete texture because certain regions of the face are occluded, as illustrated in Figure \ref{fig:overview}. 

% We project the image onto the UV space 

%We need to complete these textures as done in \cite{yamaguchi2018high, deng2018uv}.

%We consider an image-to-image translation model, that takes an incomplete texture and estimates its completed version. Similarly to \cite{deng2018uv}, we feed the incomplete texture with its mirrored  version (on vertical axis) as input to the network. To train this model, we create a dataset that simulates incomplete textures by rendering faces at random camera angles, then projecting it back to UV space - this creates occlusions and non-visible areas similarly to real images. Furthermore, to increase generalization to in-the-wild images, we perform data augmentation by shading the images (in UV space) with random lighting conditions.
%We train this network by minimizing the L2 loss between the estimated completed texture and the ground truth.

Similarly to \cite{deng2018uv}, we estimate a complete texture map from an incomplete map alongside its mirrored version. To train this model, we create synthetic textures simulating in-the-wild settings from light stage data. We render faces from random camera angles and HDR maps, then project onto UV space. This creates self-occlusions and non-visible areas similar to real images. We train an image-to-image translation network, by minimizing the $\ell_1$-distance between the estimated texture and the ground truth.

% We then re-project onto UV space  

%To complete this texture, we first concatenate (on channel wise) this texture with it's  mirrored version (on vertical axis) and then forward the resultant texture to a texture completion U-net network. To efficiently train this network and ensure generalization for in-the-wild setup, we perform the following data augmentation. For every subject in our light stage data, we render an  image under random viewing, expression and lighting conditions and project back the rendered image to uv space. We repeat this for every subject in the dataset. We generate a set of 10000 incomplete textures with their corresponding complete texture. We train this network by minimizing the L2 loss between the generated texture and the GT one. More details on the data augmentation used can be found in the supp. material (\ad{sec}).        

\subsection{Light normalization}
\label{ssec:method_ln}
\begin{figure}
    \centering
    \includegraphics[width=\linewidth]{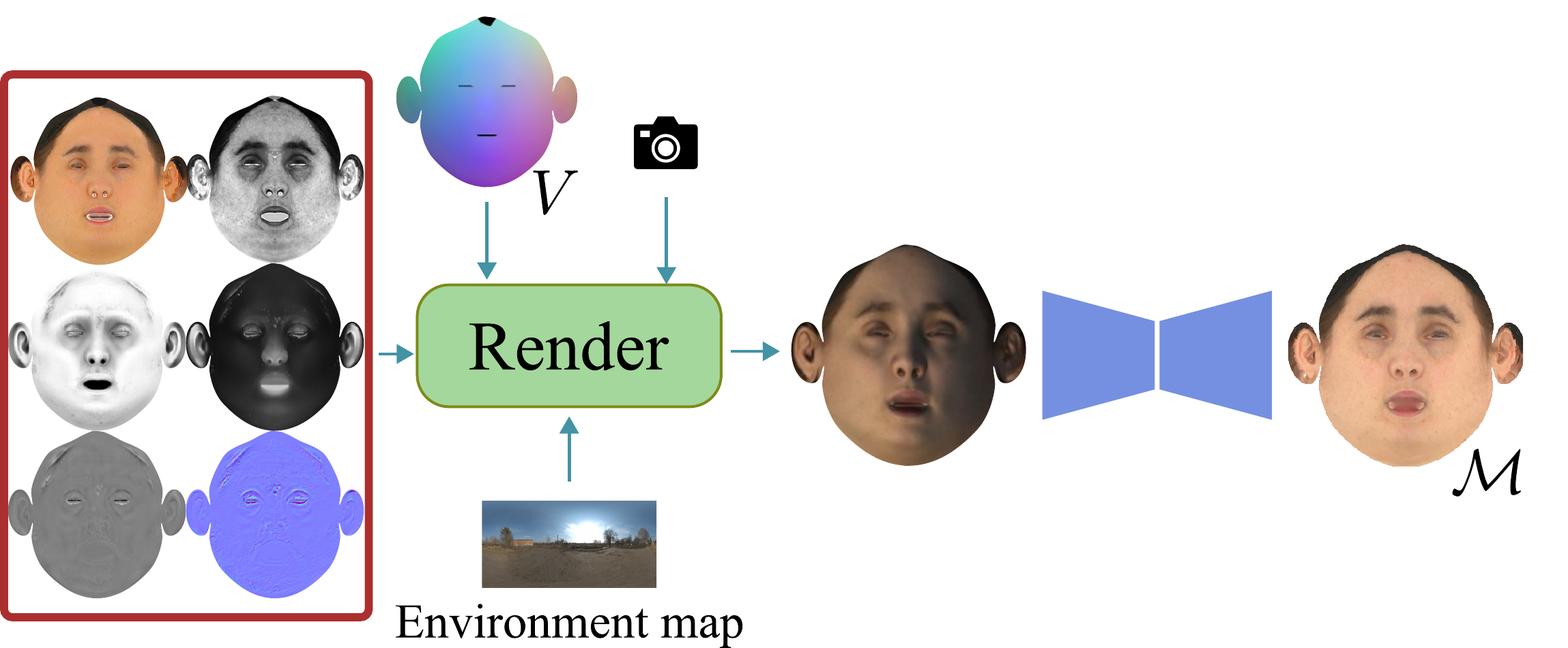}
    \caption{Light normalization step: We render faces in UV space with random light and camera positions. The network then estimates a light-normalized map $\mathcal{M}$.}
    \label{fig:ln_diagram}
\end{figure}

%but modeling ambient light with point lights provides only a crude approximation.
%Images collected in-the-wild often present challenging light conditions (e.g. strong directional light), that are hard to model with simple lighting models. Previous work \cite{lattas2021avatarme++} use a set of point lights instead that provides a very rough approximation of real light and cannot capture global and indirect illumination like Spherical Harmonics \cite{ramamoorthi2001relationship, sloan2023precomputed}. \lgh{We consider} a light normalization stage that produce images with uniform light on the face (i.e. mimic the controlled light conditions of the light stage). This stage removes strong directional shadows, enabling the effective use of SH lighting, and avoiding baking light in the estimated reflectance maps.

The light normalization step aims to produce uniform light on the subject, mimicking the controlled light conditions of the light stage.
This enables using simple light models for the remaining of the pipeline, such as Spherical Harmonics \cite{ramamoorthi2001relationship, sloan2023precomputed}, even for in-the-wild images, that can present complex lighting conditions. If this step is omitted, unexplained light contributions (e.g. strong directional lights) end up baked in the final estimated reflectance maps.

Figure \ref{fig:ln_diagram} illustrates the training procedure for the light normalization network. We first perform the shading in UV space, using the mesh geometry $V$ and its associated texture maps.
%: diffuse $\mathcal{D}$, specular $\mathcal{S}$, ambient occlusion $\mathcal{A}$, translucency $\mathcal{T}$ and normal map $\mathcal{N}$. 
We use random environment maps, camera positions
%(projected onto the \lgh{tangent space}) 
with the rendering Equation \ref{eq:render}. The resulting texture is fed to a U-Net architecture that generates a normalized texture $\mathcal{M}$. The network is trained to minimize the $\ell_1$-distance between the estimated normalized texture and the target ground truth. The target texture is obtained by projecting the images from all light stage camera views onto UV space, then aggregating them in a single image.
%The network is trained to minimize the $\ell_1$-distance between the estimated normalized texture and the ground truth obtained by the aggregation of all the light stage camera views in the UV space. 

%We generate a set of 10K images for training.

%We then relight this texture using the corresponding geometry with random HDR environment map, selected out of 200 HDR maps.  We perform relighting in local tangent space by projecting the camera position to that space. We use Cook-torrance BRDF to model diffuse and specular effects. We generate a set of 10K images for training. The network used here is U-net based architecture. 

%\ad{I think it is important to showcase some images on this. We may want also to provide more details on the local shading (it is the same as the one used in the next section}

%We start from the texture map obtained from the Multi View Stereo (MVS)stage. This texture contains the aggregation of all the camera views in the same canonical uv space. We then relight this texture using the corresponding geometry and from a random HDR environment map, selected out of 200 HDR maps.  For this, we apply local shading in the tangent space. We use Cook-torrance BRDF to modle diffuse and specular 
\subsection{Intrinsic texture maps estimation}
\label{ssec:method_td}

% usid in a Differentiable Shading

\begin{figure}
    \centering
    \includegraphics[width=\linewidth]{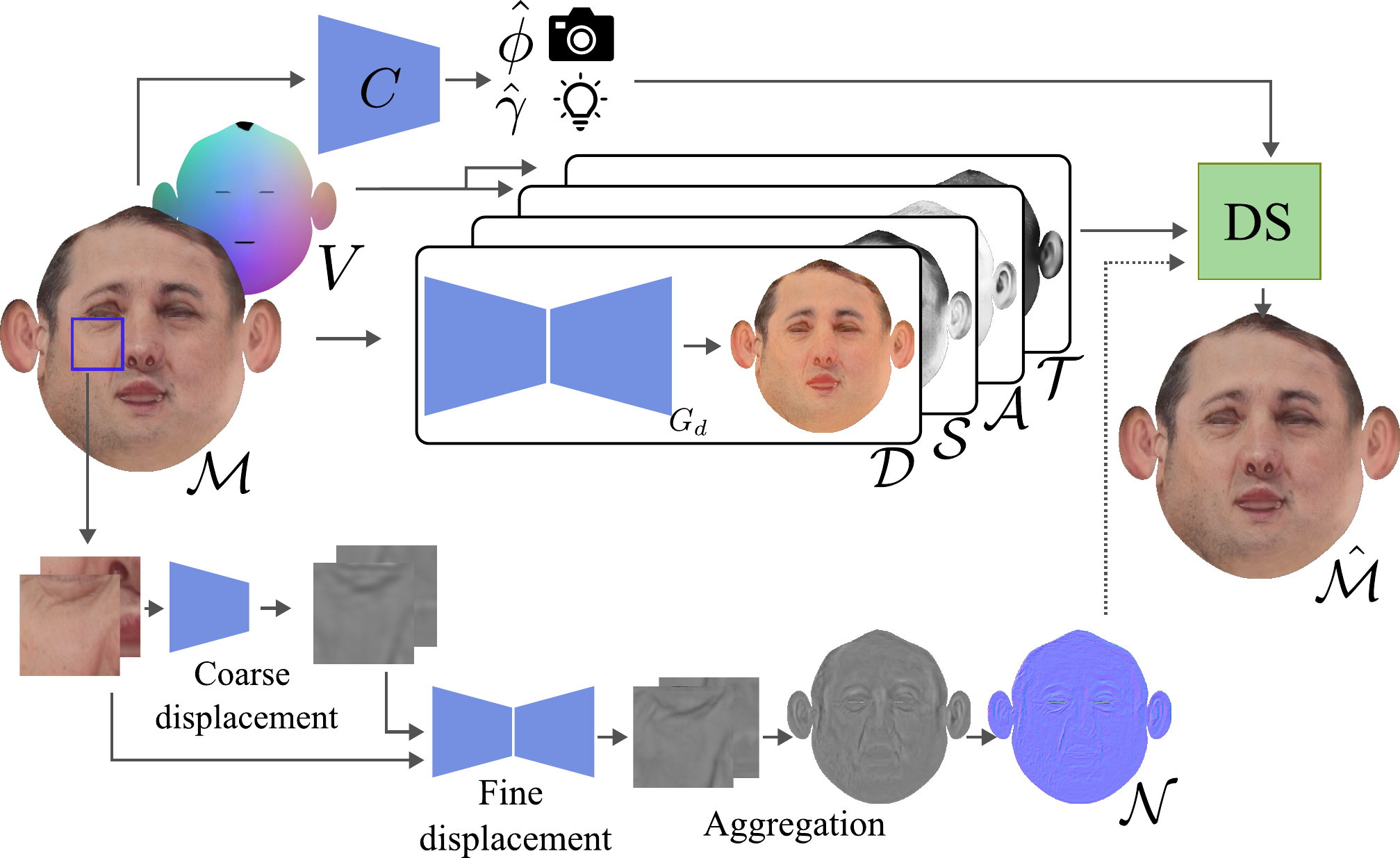}
    \caption{Texture estimation pipeline: Reflectance and normal maps are estimated by reconstructing the light-normalized texture $\mathcal{M}$ using Differentiable Shading (DS).}
    \label{fig:texture_estimation}
\end{figure}

The normalized texture $\mathcal{M}$ and the estimated geometry $V$ are used to estimate different face attribute maps:  Diffuse $\mathcal{D}$, Specular $\mathcal{S}$, Ambient Occlusion $\mathcal{A}$, Translucency $\mathcal{T}$ and Normal map $\mathcal{N}$, as depicted in Figure \ref{fig:texture_estimation}.

A separate network is used to estimate each of the reflectance maps. This architecture could be trained using only light stage data by supervising the output of each network. To achieve better generalization outside the light stage, we seek training on in-the-wild images as well. For this, we add an encoder $C$ to estimate the residual light $\hat{\gamma}$ and the camera position $\hat{\phi}$, and introduce a new  differentiable shading formulation (DS in Figure \ref{fig:texture_estimation}). DS  outputs a new shaded texture $\hat{\mathcal{M}}$ which can then be compared to $\mathcal{M}$. This allows our model to be trained in a semi-supervised manner, combining light stage and in-the-wild datasets. The encoder $C$ captures residual light that was not removed during the Light Normalization step (Section \ref{ssec:method_ln}). This prevents the model from baking this light into one of the texture maps.

% \my{} we seek training on in-the-wild images as well  -> just to note that "in the wild" is generally not italicized anywhere else, so why here?
% which can be then compared
% This prevents the training to bake this light in one of the texture maps.

To estimate the normal map $\mathcal{N}$, we first predict the displacement map following a patch-based approach as in \cite{chen2019photo}.
%in \cite{cao2015real} and we use 2 networks to estimate the displacement map as in \cite{chen2019photo}. 
First, a PCA basis is built on patches of displacement maps from the light stage data. A coarse displacement encoder (refer to Figure \ref{fig:texture_estimation}), estimates the PCA coefficients that best reproduce a patch of the texture $\mathcal{M}$. This PCA captures repeated patterns (e.g. wrinkles) and low to medium frequency details. This approximation is further refined by a second U-Net model that complements the patch with high frequency details (e.g. skin pores and moles). Finally, the normal map is obtained from the displacement map using a Fast Fourier convolution. 

% \my{} For A and translucency T -> what is A?  you named all the others...

To estimate the remaining maps ($\mathcal{D}, \mathcal{S}, \mathcal{A}, \mathcal{T}$), %inspired by \cite{lattas2021avatarme++}, 
we use image-to-image translation networks. 
%However, we propose a novel semi-supervised training scheme. 
For the diffuse $\mathcal{D}$ and specular $\mathcal{S}$, the generators predict the maps from the normalized texture image $\mathcal{M}$. For $\mathcal{A}$ and translucency $\mathcal{T}$, the generators also use the geometry $V$ projected onto UV space.  We use adversarial training, by defining a discriminator for each texture map. These networks are trained jointly, with the following loss:

\begin{align}
 \label{eq:loss_tex}
\mathcal{L} = L_\text{shading} + \sum_{ \mathcal{D}, \mathcal{S}, \mathcal{A}, \mathcal{T} }{\left( L_\text{GAN}(G, D) + L_\text{sup} \right) },
\end{align}
where $L_\text{shading}$ is the $\ell_1$ reconstruction loss between $\mathcal{M}$ and $\hat{\mathcal{M}}$, $L_\text{GAN}(G, D)$ is the adversarial loss described in \cite{isola2017image}, applied for each texture map separately. Finally, when training with light stage data, we also compute $L_\text{sup}$ as the $\ell_1$ loss between the ground truth and predicted maps. 
We omitted weighting coefficients ($\lambda$) of the loss term for better readability. They are specified in the supplementary material.

\paragraph{Differentiable shading.}
The differentiable shading \emph{DS} (in Figure \ref{fig:texture_estimation}), takes the predicted textures ($\mathcal{D}$, $\mathcal{S}$, $\mathcal{A}$,  $\mathcal{T}$ and  $\mathcal{N}$), camera parameters $\hat{\phi}$, light coefficients $\hat{\gamma}$ and outputs a new texture map $\hat{\mathcal{M}}$. 
%To calculate the final surface color 
%$\hat{m}^i$ at every pixel in $\hat{\mathcal{M}}$, 
We formulate our shading equation, for each pixel $\hat{m}$ in $\hat{\mathcal{M}}$, extending the Microfacet BRDF model \cite{torrance1967theory} as follows: 
%The final color $\hat{m_i}$ (for each pixel  $\hat{\mathcal{M}}$) is computed by aggregating diffuse, specular and sub-surface scattering contributions. For readability, we show separate equations for each: $\mathcal{B}_d^i$, $\mathcal{B}_{sss}$ and $\mathcal{B}_s^i$, respectively
%, extending the Cook-Torrance BRDF as follows: 
% \begin{align}
%     \hat{m_i} = (1 - s_i) \cdot (\mathcal{B}_d^i + \mathcal{B}_{sss}^i) +  s_i \cdot \mathcal{B}_s^i,
% \label{eq:render}
% \end{align}
\begin{align}
    \hat{m} = \mathcal{B}_d + \mathcal{B}_{sss} +  \mathcal{B}_s,
\label{eq:render}
\end{align}
where $\mathcal{B}_d$, $\mathcal{B}_{sss}$ and $\mathcal{B}_s$ are the diffuse, subsurface scattering and specular contribution respectively, described in the equations below.  
%The weighting factor $c_i, r_i \in\mathbb{R}$ are the diffuse and specular intensity respectively, at pixel $i$ obtained from the diffuse $\mathcal{D}$ and specular albedo map $\mathcal{S}$.

To model the diffuse component of the skin $\mathcal{B}_d$, %and inspired by \cite{gotardo2018practical}, 
we extend the diffuse Lambertian model with an additional term $a$ that captures residual diffuse ambient occlusion in areas that receive less light such as wrinkles, folds and nostrils. $\mathcal{B}_d$ is calculated as follows:
\begin{equation}
    \mathcal{B}_d = d \cdot  a \cdot \sum_{l = 0}^{2} \sum_{m = -l}^{l} A_{l} \cdot \hat{\gamma}^m_{l} \cdot Y^m_{l}(n),
    \label{eq:diffuse}
\end{equation}
where $d$, is the diffuse albedo obtained from the diffuse $\mathcal{D}$ map, and $a$ is the residual ambient occlusion term  from $\mathcal{A}$. $A_{l}$ are the Lambertian BRDF coefficients \cite{Ramamoorthi2001efficient, ramamoorthi2001relationship}. $\hat{\gamma}$ are the SH coefficients, and $Y^m_{l}$ are the SH basis \cite{ramamoorthi2001relationship}. $n$ is the pixel normal (using normal mapping). For the diffuse component, we use 3\textsuperscript{rd}-order spherical harmonics ($l = 2$)  as it has been shown in \cite{Ramamoorthi2001efficient} that it captures 99\% of the reflected radiance.

Next, we introduce a new subsurface scattering contribution $\mathcal{B}_{sss}$ which indicates where thinner parts of the face (e.g. nose, lips and ears) scatter more light than other thicker areas, defined as:
%Inspired by \cite{barre2011approximating}, we introduce a new term defined as: 
\begin{align}
    \mathcal{B}_{sss} = d \cdot \sum_{l = 1}^{2} \sum_{m = -l}^{l} S_{l} \cdot \hat{\gamma}^m_{l} \cdot Y^m_{l}(n),
    \label{eq:sss}
\end{align}
with $S_{l} = e^{- l^2 / t^4}$, with ${t}$ being the translucency value from $\mathcal{T}$. Intuitively, thicker parts of the face would have a low value in $\mathcal{T}$ which in turn would negate the $\mathcal{B}_{sss}$ contribution through $S_l$ nearing $0$. %\ad{It is noted that in equation \ref{eq:sss}, the first band ($l = 0$)  which has a constant contribution on the the whole geometry \cite{Ramamoorthi2001efficient} is excluded, as we want to model local scattering effect. We found this to give a good approximation of the scattering as we show in the supp. material. }

Finally, we compute the specular component $\mathcal{B}_s$ as the spatial convolution of the SH light representation with the BRDF roughness kernel. This kernel is constant in the simplified Microfacet BRDF model that we use. Thus our specular contribution $\mathcal{B}_s$ is defined as:
\begin{align}
    \mathcal{B}_s = f \cdot \sum_{l = 0}^{8} \sum_{m = -l}^{l} R_{l} \cdot \hat{\gamma}^m_{l} \cdot Y^m_{l}(r),
    \label{eq:specular}
\end{align}
where $f = s + (1 - s)(1 - \cos{\theta})^5 $  is the Fresnel reflection \cite{walter2007Microfacet} calculated using Schlick's approximation \cite{schlick1994inexpensive}. $f$ quantifies the proportion of light that is reflected and depends on the angle at which light hits the surface. $\theta$ is approximated with the angle between the normal vector and the camera view direction \cite{ramamoorthi2001signal}.  $s \in\mathbb{R}$ is specular intensity from $\mathcal{S}$.
%and depends on the angle at which light hits the surface. We follow the same approximation as  in \cite{ramamoorthi2001signal}, this angle is very close the the angle between the viewing angle and the normal vector (known as the angle of incidence, denoted as "theta") and the material's specular  $s_i \in\mathbb{R}$ obtained from the specular albedo map $\mathcal{S}$.
%where $f_i = s_i + (1 - s_i)(1 - \cos{\theta} $ is the Schlick approximation \cite{schlick1994inexpensive} of the Fresnel term  \cite{walter2007Microfacet} that represent the amount of reflected light based on the angle of incidence $\theta$ between the normal vector and the viewing direction and the specular property $s_i \in\mathbb{R}$ of the material  obtained from the specular albedo map $\mathcal{S}$.
%where $s_i \in\mathbb{R}$ is specular intensity at pixel $i$ obtained from the specular albedo map $\mathcal{S}$.  
$r_i$ is the reflection vector of the viewing vector (obtained from the camera position $\hat{\phi}$) according to the surface normal \cite{walter2007Microfacet}, and $R_{l}$ are the SH coefficients of the BRDF function corresponding to the roughness \cite{ramamoorthi2001signal, mahajan2007theory}. For the specular contribution, we use 9\textsuperscript{th}-order order spherical harmonics ($l = 8$) for better approximation of the specular reflection \cite{dib2021towards}.

Equation \ref{eq:render} is fully differentiable with respect to all texture maps, allowing for gradient-based optimization of the generators. 
We note that the diffuse $d$ and ambient occlusion $a$ have the same contribution, and therefore cannot be separated using only self-supervised training. For instance, the model can bake all the information in a single diffuse map which results in a sub-optimal separation. Our semi-supervised training scheme (Equation \ref{eq:loss_tex}) allows for this separation, by leveraging ground truth data from light stage in conjunction with the unlabeled data. 

% in conjuction to the unlabeled data

%For instance, the model can bake all the information in a single diffuse map which results in a less control over these parameters. As we seek disentangled attributes estimation \textit{in-the-wild}, our semi-supervised training scheme (eq. \ref{eq:loss_tex}), leverage ground truth data from light stage and unlabeled data to achieve a faithful separation between these attributes.}
\subsection{Super-resolution}
We use super-resolution at the last step, to obtain 4K texture maps. 
%We train an ESRGAN model \cite{wang2021real}, with pairs of 4K textures and low-resolution textures ($256 \times 256$), obtained by downsampling the textures from the light stage dataset. We use two networks, one that upsamples the input to 1K and the other one goes for 1K to 4K, as in \cite{li2020learning}.
We train two ESRGAN models \cite{wang2021real}: one that upsamples the input to 1K and one  that upsamples to 4K, as in \cite{li2020learning}. Both are trained with data from the light stage.

\section{Experimental protocol}
\label{sec:exp_protocol}

\paragraph{Datasets}
%Our light stage is composed of 54 synchronized cameras (spot lights nb?) that capture the face face under 360 degrees. 
The light stage dataset is composed of 890 subjects, captured with a neutral pose. For every subject, 12 cameras capture frontal and side views. The geometry is obtained using a Multi-View reconstruction pipeline, followed by registration to a standard topology. 
Each subject has diffuse, specular, displacement, ambient occlusion and translucency maps in 4K.
% capture frontal and side views of the subject.
%Each subject has the diffuse, and specular texture maps at $4K$. The displacement, ambient occlusion and translucency maps are obtained from the high resolution mesh using commercial software\footnote{\url{https://marmoset.co/toolbag/}}.
We keep 50 subjects for validation, and use the remaining for training.
%For every subject, the diffuse and specular albedos are extracted using Debevec~\textit{et al.}  \cite{debevec2000acquiring}. The mesh is obtained using MVS (cite realityCapture?) followed by a registration and cleanup process to get the final mesh in our topology. Displacement, ambient occlusion and translucency maps are obtained from the high resolution mesh (the output of MVS). We keep 50 subjects for validation and adjusting the hyper parameters of the networks. 
%
%\paragraph{In-the-wild Datasets}
%For training the 3D reconstruction model, we use the FFHQ dataset \cite{karras2019style}, composed of 70k images. We crop and resize the input images to the resolution of $256 \times 256$, we apply standard data augmentation (brightness, rotate, scale and flip).

We train the 3D reconstruction model with the light stage dataset and 70k images from the FFHQ dataset \cite{karras2019style}. We crop and resize the images to the resolution of $512 \times 512$, we apply standard data augmentation (brightness, rotation, scale and flip).
For texture estimation, we use the light stage dataset and FFHQ-UV \cite{Bai_2023_CVPR}, that contains 54k textures obtained from StyleGAN \cite{Karras2019stylegan2}. We normalize the FFHQ-UV textures using our light normalization network (Section \ref{ssec:method_ln}). The textures are processed in resolution $512 \times 512$.

\paragraph{Training}
%For the 3D geometry estimation (Sec. \ref{ssec:method_shape}), when training $\mathcal{E}$, we take a batch consisting of 32 samples, which we divide equally between labeled (light stage) and unlabeled (in-the-wild) data. 
%\ma{For the 3D geometry estimation (Sec. \ref{ssec:method_shape}), we train $\mathcal{E}$ with a batch size of 32 equally divided between light stage and in-the-wild data.}

%For $\mathcal{E}$, a Resnet-50 \cite{he2016identity} is used. For $\mathcal{F}$ we use the GNN architecture from \cite{aliari2023face}. We train $\mathcal{E}$ and $\mathcal{F}$ for 20 epochs by minimizing the loss functions in equations \ref{eq:loss_geom} and \ref{eq:loss_geom_sup}.

For the 3D geometry estimation (Section \ref{ssec:method_shape}), $\mathcal{E}$ is a ResNet-50 \cite{he2016identity} while for $\mathcal{F}$ we use the same architecture as \cite{aliari2023face} from their open-source implementation. We train both networks for 20 epochs, using batch size of 32 images, equally divided between light stage and in-the-wild data.

%For $\mathcal{E}$, a Resnet-50 \cite{he2016identity} is used. For $\mathcal{F}$ we use the GNN architecture from \cite{aliari2023face}. We train $\mathcal{E}$ and $\mathcal{F}$ for 20 epochs by minimizing the loss functions in equations \ref{eq:loss_geom} and \ref{eq:loss_geom_sup}.

The texture completion network is trained for 20 epochs on the light stage data following the data augmentation described in Section \ref{ssec:method_tc}. We train the light normalization network for 20 epochs, using random environment maps sampled from a collection of 190 HDR maps\footnote{\url{https://polyhaven.com}}. 

The displacement map estimation is done in patches in the full resolution (4K). We manually selected 20 wrinkled subjects from the training set to obtain the PCA basis for coarse displacement, as done in \cite{chen2019photo}. After training the coarse displacement encoder for 5 epochs, we jointly train the coarse and fine displacement networks for 15 epochs.

%To build the PCA for the coarse displacement estimation, following the observation from \cite{chen2019photo}, we manually selected 20 subjects from the training set with visible wrinkles. We warm-start the coarse displacement encoder for 5 epochs and then we jointly train the encoder and the fine displacement network for 15 epochs.

The texture estimation networks are trained on the light stage data and the FFHQ-UV dataset. We use a batch size of 2 containing a labeled and an unlabeled image. 
All the generators are 5 layer U-Nets with skip connections \cite{ronneberger2015u}. We use the patch GAN multi-scale discriminators from \cite{park2019SPADE}. 

%\commentlgh{Number of epochs and time it takes to train}
%it's important to note that we noticed that when using a higher regularization term for the landmarks $\lambda_{land}$,  spikes  appear on the mesh at the landmark positions, which, necessitates the use of a higher value for the Laplacian term. This adjustment tends to over-smooth out the mesh. On the other hand, removing the landmarks loss leads  the model not to converge. %% LGH: this sound strange - we may need 
%For the texture estimation, we first trained the displacement map networks separately, followed by training the reflectance estimation models, since we ran into out-of-memory issues when training jointly with the reflectance maps.
%\ad{i am not sure if we need to say that we run out of memory. that looks like we do not have enough gpu or it revails our incapacity to run on multi gpu :(}
% \lambda_{ph} \cdot L_\text{photo} + \lambda_{land} \cdot L_\text{landmark} +  \lambda_{lap} 
\section{Results}
\label{sec:results}
%5. Results
%   5.1 3D Reconstruction (comparison on REALY, visual comparison to HIFACE, etc, 
%     5.1.x Ablations on linear vs non linear model, laplacian smoothing)
%   5.2 Full-avatar reconstruction (user study, visual comparison to Fitme, Relightify, 
%     5.2.x Ablations on light normalization, AO/THK maps

\subsection{Geometry estimation}

In this section, we evaluate our method for 3D shape reconstruction. We compare to different state-of-the-art methods, including methods that specialize in geometry prediction without supporting texture map generation. 
%We compare the produced meshes in terms of reconstruction accuracy, visual similarity to the reference picture and overall perceived visual quality.

\begin{figure}
    \centering
    \includegraphics[width=\linewidth]{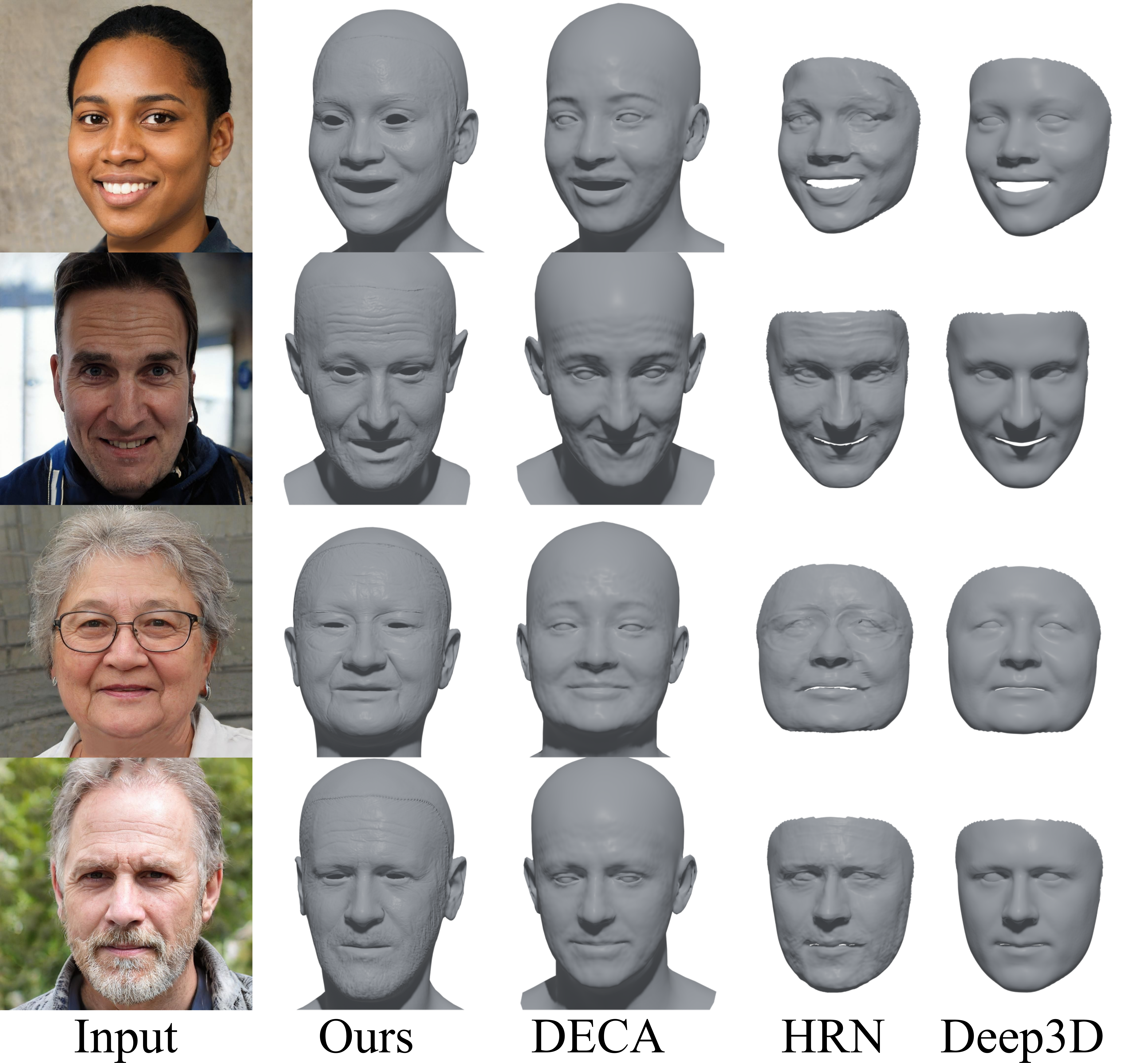}
    \caption{Comparison of 3D face reconstruction methods.} 
    \label{fig:geometry_comparison}
\end{figure}

%First we visually compare results from our method to DECA \cite{feng2021learning}, HRN \cite{lei2023hierarchical} and Deep3D \cite{deng2019accurate}. 
%HiFace \cite{chai2023hiface}and Lee~\textit{et al.}\cite{lee2020uncertainty}. HiFace, DECA, HRN and Deep3D uses a linear morphable model (PCA) while Lee~\textit{et al.}\cite{lee2020uncertainty} use a non linear morphable model. 
%Figure \ref{fig:geometry_comparison} shows renders for the estimated geometry of each method, under the same rendering conditions. Our method produces meshes with more similar likeness to the input images, notably on the nose, eyes, cheek and the overall shape of the face.
%It also captures higher-frequency details (such as wrinkles and folds) with more fidelity than the other methods, resulting in a  more expressive reconstruction. The details captured by DECA are barely visible or sometimes missed (last two subjects). Deep3D restricts the estimation to a 3DMM and does not capture any of the folds and wrinkles. The mesh obtained by HRN have noticeable artefacts on the eyes and nose. Comparison on more subjects can be found in the supplementary material.

First we visually compare our results to 3 available state-of-the-art methods: DECA \cite{feng2021learning}, HRN \cite{lei2023hierarchical} and Deep3D \cite{deng2019accurate}. 
Figure \ref{fig:geometry_comparison} shows the estimated geometry of each method under the same rendering conditions. Our method produces meshes with more similar likeness to the input images, notably on the nose, eyes, cheeks and the shape of the face.
It also better captures higher-frequency details such as wrinkles and folds than the other methods. While HRN captures some of the details, they are not as fine and precise as ours. Moreover, HRN produces noticeable artifacts around the eyes and nose. The details captured by DECA are not as visible and sometimes missing. Deep3D does not predict displacement maps, and does not capture any of the folds and wrinkles. However, DECA and Deep3D avoid baking glasses in the geometry as HRN and our method do. The supplementary material contains additional comparisons.

%
%Figure \ref{fig:geometry_comparison} compares the geometry we obtain with DECA \cite{feng2021learning}, HRN \cite{lei2023hierarchical} and Deep3D \cite{deng2019accurate}\footnote{Comparison against HiFace \cite{chai2023hiface} and \cite{lee2020uncertainty} are pending}. HiFace, DECA, HRN and Deep3D uses a linear morphable model (PCA) while \cite{lee2020uncertainty} use a non linear morphable model similar to us. Deca and HRN 
%\lgh{
%Figure \ref{fig:geometry_comparison} compares the geometry we obtain with DECA \cite{feng2021learning}, HRN \cite{lei2023hierarchical} and Deep3D \cite{deng2019accurate}. Our method produces meshes with more similar likeness to the input images, notably on the nose and the overall shape of the face. It also captures higher-frequency details (such as wrinkles) with more fidelity. On the other hand, we notice on line 3 that eyeglasses are incorrectly captured in the geometry, which also occurs in HRN, that also estimates high-frequency details.}
\begin{table}[]
\centering
\begin{adjustbox}{max width=\linewidth}
\begin{tabular}{l|c|c|c|c|c} 
\multicolumn{1}{c}{} & \multicolumn{5}{c}{Error (mm)} \\

Method         & Nose                           & Mouth                          & Forehead                       & Cheeks                         & All   \\ \midrule
MICA \cite{MICA:ECCV2022}      &  1.585 $\pm$ 0.325 & 3.478 $\pm$ 1.204  & 2.374 $\pm$ 0.683 & 1.099  $\pm$ 0.324 &  2.134 \\
%EMOCA \cite{EMOCA:CVPR:2021}      & 1.868  $\pm$ 0.387 & 2.679  $\pm$ 1.112  & 2.426 $\pm$ 0.641 &  1.438 $\pm$ 0.501 & 2.103 \\
%SynergyNet \cite{wu2021synergy}      & 2.026	  $\pm$ 0.532 &  1.731 $\pm$ 0.502  & 2.679 $\pm$ 0.741 & 1.647  $\pm$ 0.622 & 2.021 \\
DECA \cite{feng2021learning}      &  1.697 $\pm$ 0.355 & 2.516 $\pm$  0.839 & 2.394	 $\pm$ 0.576 &  1.479 $\pm$ 0.535 & 2.010 \\
%3DDFA-v2 \cite{guo2020towards}      & 1.791  $\pm$ 0.542 & 1.591 $\pm$ 0.488  & 2.413 $\pm$ 0.537 & 1.856  $\pm$ 0.701 & 1.913 \\
PSL \cite{PSL}    &  1.708 $\pm$ 0.349 & 1.708 $\pm$ 0.563  & 2.350  $\pm$ 0.551 & 1.593 $\pm$ 0.540 & 1.882 \\
%GanFit \cite{gecer2019ganfit}     & 1.928  $\pm$ 0.490  & 1.812  $\pm$  0.544 & 2.402 $\pm$ 0.545 &  1.329 $\pm$ 0.504 & 1.868 \\
%MGCNet \cite{shang2020self}    &  1.771 $\pm$ 0.380  & 1.417 $\pm$  0.409 & 2.268 $\pm$ 0.503  & 1.639  $\pm$ 0.650 & 1.774 \\
AlbedoGan \cite{rai2023towards}     & 1.656  $\pm$ 0.374 & 2.087 $\pm$ 0.839  & 2.102 $\pm$ 0.512 & 1.141  $\pm$ 0.303 & 1.746 \\
Deep3D \cite{deng2019accurate}    & 1.719  $\pm$ 0.354 & 1.368 $\pm$ 0.439  & 2.015 $\pm$ 0.449  & 1.528  $\pm$ 0.501  & 1.657  \\
HRN \cite{lei2023hierarchical}    & 1.722 $\pm$ 0.330 & 1.357 $\pm$ 0.523  & 1.995 $\pm$ 0.476 & 1.072  $\pm$0.333 & 1.537 \\
HiFace (w/o synthetic) \cite{chai2023hiface}     & 1.227 $\pm$ 0.407 & 1.787 $\pm$ 0.439  & 1.454 $\pm$ 0.382 & 1.762  $\pm$ 0.436 & 1.558 \\ 

HiFace \cite{chai2023hiface}     & 1.036 $\pm$ 0.280 & 1.450 $\pm$ 0.413  & 1.324 $\pm$ 0.334 & 1.291  $\pm$ 0.362 & 1.275 \\ \midrule

Baseline (linear 3DMM) & 1.815 $\pm$ 0.516 & 1.725 $\pm$ 0.576 & 2.550 $\pm$ 0.825 & 1.469 $\pm$ 0.511 & 1.890 \\ 
MoSAR (ours)      & 1.499 $\pm$ 0.366 & 1.424 $\pm$ 0.462 & 1.950 $\pm$ 0.559 & 1.128 $\pm$ 0.303 & 1.500 \\ 
\bottomrule

\end{tabular}
\end{adjustbox}
\caption{Reconstruction errors on the REALY benchmark.}
\label{tab:REALY}
\end{table}

For a quantitative analysis, we compare our method to state-of-the-art methods using REALY \cite{REALY}, a publicly available benchmark. This evaluation framework uses multi-view rendered portrait images of 100 high quality scans from the HeadSpace dataset \cite{dai2020statistical}. Geometric reconstruction errors are computed separately on different face regions which reduces imprecision caused by alignment errors and allows for a fine-grained analysis of the results. For each face region, the mean of the per vertex $\ell_2$-distance between the predictions and the ground truth is reported in Table~\ref{tab:REALY}.
% uses multi-view rendered portraits image
%Table~\ref{tab:REALY} reports the results of the best performing methods available on the REALY website\footnote{\url{https://realy3dface.com/}} at the time of writing this paper. 
%Table~\ref{tab:REALY} reports the results.
%Our method achieves second place on average reconstruction error, being surpassed only by the full HiFace model \cite{chai2023hiface}, that was trained with 200k synthetic images with ground-truth meshes. We notice that this synthetic dataset brings significant improvement to HiFace, which could also be applied to our method. 

Our method ranks second on reconstruction error averaged for all parts, being surpassed only by the full HiFace model \cite{chai2023hiface} when trained with 200k synthetic images with ground-truth meshes. The addition of the synthetic dataset significantly improves HiFace's results, which otherwise would perform a bit worst than ours. We believe that our method would similarly benefit from this additional data.

Our results indicate that non-linear morphable models can perform competitively compared to the dominating linear 3DMM. We included in Table~\ref{tab:REALY} an implementation of our method using a linear 3DMM, a PCA created with the same light stage dataset. Except for the geometry model, all other factors were kept constant (i.e. dataset, encoder architecture).
These results show that the improvement stems from the higher expressivity of non-linear modeling. 

\subsection{Texture estimation ablation}
\label{sec:ablation}
This ablation study evaluates the effect of light normalization (Section \ref{ssec:method_ln}) and the proposed semi-supervised training scheme (Section \ref{ssec:method_td}) on the quality of the predicted texture maps. 
As a baseline, we train our system in a fully-supervised manner without light normalization, using only light stage data. This means that our differential shading is not required for self-supervision, thus, we ignore the term $L_\text{shading}$ on Equation \ref{eq:loss_tex}. 
%This is denoted as \emph{Supervised} in table \ref{tab:ssim}. 
%Note that for this model differential shader is not used. 
Then, we include the light normalization to the baseline model to measure its impact. For completeness, we train a model in a purely self-supervised manner, without any light stage data.

To evaluate these systems on settings similar to in-the-wild, while having access to ground truth maps, we render subjects from our light stage validation data under novel lighting conditions using random HDR maps. From 50 subjects, we generate 1000 renders. We report the Structural Similarity Index (SSIM) \cite{wang2004image} between the predicted and ground-truth maps.
\begin{figure*}
    \centering
    \includegraphics[width=\linewidth]{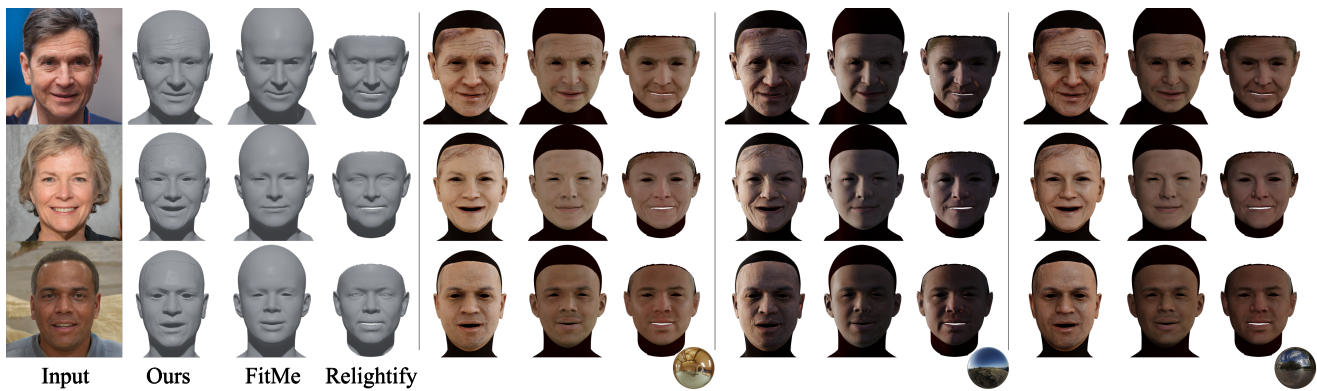}
    \caption{Comparison of the estimated geometry and renders under 3 lighting conditions, against FitMe \cite{lattas2023fitme} and Relightify \cite{papantoniou2023relightify} }
    \label{fig:vsLattas}
\end{figure*}
% 3 light conditions
% \usepackage{tabularray}
\begin{table}
\centering
\begin{adjustbox}{max width=\linewidth}
\begin{tabular}{c|c|c|c|c} {
}
Method     & Diffuse & Specular & AO   & Transl.  \\ \toprule
Supervised & 0.69 & 0.28 & 0.61 & 0.67 \\
+ Light Normalization & 0.80 & 0.55 & 0.82 & 0.79 \\ \midrule
Self-supervised & 0.80 & 0.19 & 0.36 & 0.65 \\  \midrule
Full (Semi-supervised)       & \textbf{0.83} &\textbf{ 0.65} & \textbf{0.85} & \textbf{0.82} \\ \bottomrule

\end{tabular}
\end{adjustbox}
\caption{SSIM metric between estimated and ground truth texture maps, for different settings of our method (higher is better).}
%\caption{SSIM for different settings of our method, for each texture map compared to ground truth (higher is better).}
\label{tab:ssim}

\end{table}

Results are shown in Table~\ref{tab:ssim}. The supervised method, trained using only light stage data underperforms, indicating poor generalization when confronted to the variety of lighting conditions in-the-wild.
Our light normalization step reduces this problem by attenuating specular highlights and strong shadows. Consequently, it improves SSIM, specially on the specular map. On the other hand, the self-supervised method, which is trained without any ground-truth textures, fails to disentangle the contributions of different maps, as expected, since they are entangled in the rendering Equation \ref{eq:render}. It bakes most of the information in the diffuse albedo map and is therefore unsuitable for relighting applications.
Our full model obtains the best SSIM. It takes advantages of high-quality light stage data to better disentangle the contributions of different maps, as opposed to the self-supervised-only version. Moreover, training from in-the-wild images exposes the model to a wide variety of lighting conditions and allows for training with a larger quantity of data. This translates to improvements on every map, but especially the specular map.
% This translates in improvements on every maps, but specially the specular map.

\begin{figure}
    \centering
    \includegraphics[width=\linewidth]{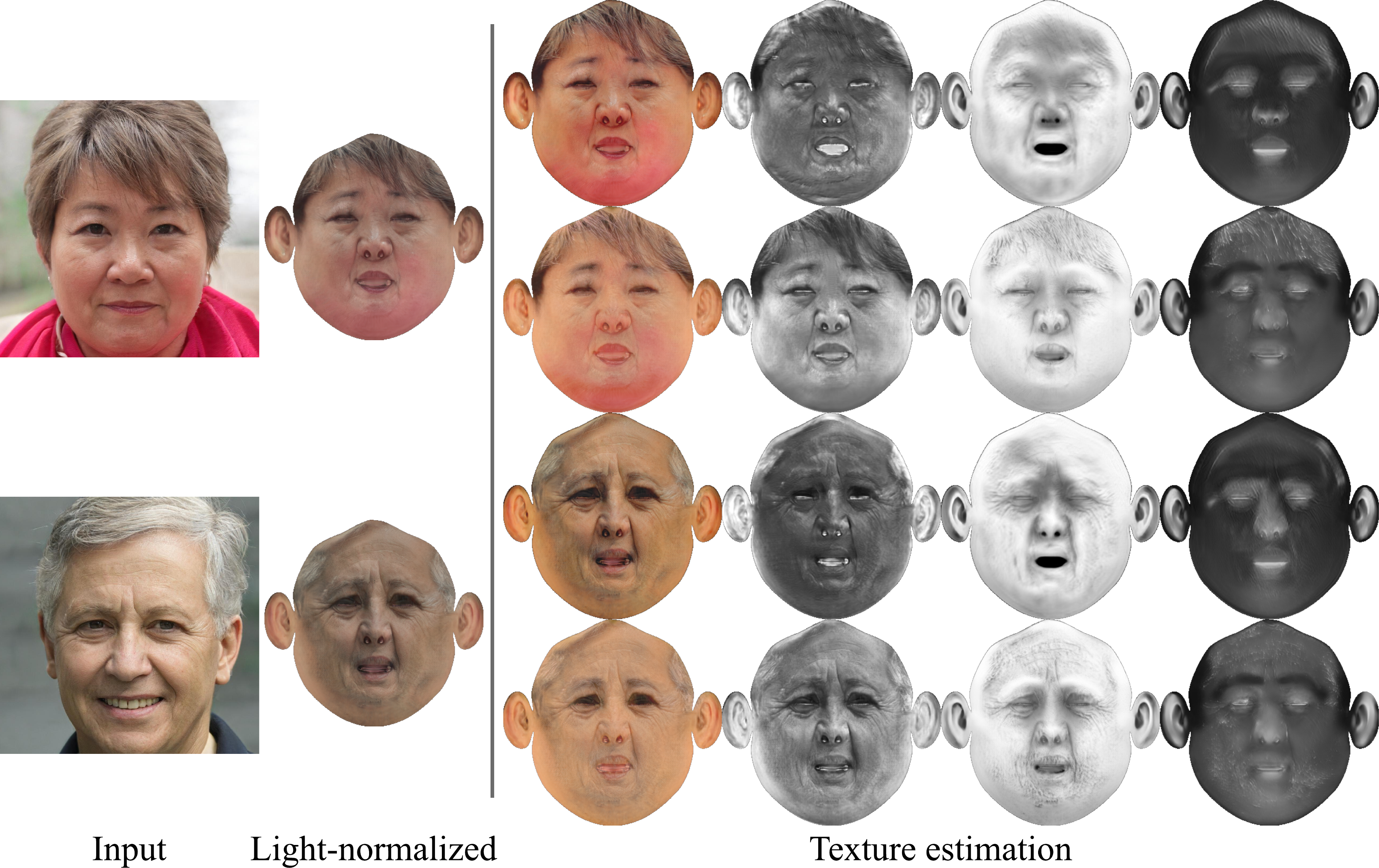}
    \caption{Reflectance estimation using the ``supervised + Light Normalization" model (first and third rows) vs the full method (second and fourth rows). }
    %The full method (semi-supervised) produces more faithful attributes separation on \textit{in-the-wild} images.}
    \label{fig:ablation}
\end{figure}

Aside from improving SSIM, our full model produces visually sharper and cleaner maps when applied to in-the-wild data, as shown in Figure \ref{fig:ablation}. Our semi-supervised training helps to better separate the intrinsic face attributes. For instance, the supervised-only model bakes specular reflections in the diffuse map. This is observed on the specular highlights of the nose, which our full model successfully removed from the diffuse map. 
The supervised model also bakes into the diffuse map both light and shadows that were not completely removed by the light normalization step. The full model correctly handles the diffuse and specular terms.
Finally, the AO map produced by the full method better captures the shading of the wrinkles and folds.

% \begin{figure}
%     \centering
%     \includegraphics[width=\linewidth]{figures/vs_fitme_relightify.pdf}
%     \caption{Comparison of the estimated geometry and a relighting of our method against FitMe \cite{lattas2023fitme} and Relightify \cite{papantoniou2023relightify} }
%     \label{fig:vsLattas}
% \end{figure}

\subsection{Avatar reconstruction}

%%%%%%%%%%%%%%%%%%%%%%%%%%%%%%%%%%%%%%%%%%%%%%%%%%%%%%
% TEXTURE MAPS EVALUATION
%%%%%%%%%%%%%%%%%%%%%%%%%%%%%%%%%%%%%%%%%%%%%%%%%%%%%%

In this section, we perform a visual comparison with the most recent state-of-the art methods that perform full-avatar reconstruction: FitMe \cite{lattas2023fitme}  and Relightify \cite{papantoniou2023relightify}. Both methods estimate geometry, diffuse, specular and normal maps from a single image. %\ad{using supervised training scheme only}. %LGH: I don't think we need to add this
Figure \ref{fig:vsLattas} shows, for each method, the estimated geometry (including normal map), and a rendering of the  avatar, under the same conditions (camera placement and environment map).
%We use the same Blender scene for the rendering.
%Our method estimates ambient occlusion and translucency. 
We focus the comparison on the skin reflectance, and therefore we removed the scalp, eyes and mouth interior of FitMe and Relightify, since our method do not estimate them. We refer the reader to the supplementary material for the comparison with more subjects and lighting conditions.

%Our method has a better face and expression estimation, as well as the fine details (such as wrinkles and folds). Notably, the shape of the nose is more accurate than FitMe for most subjects for instance. The fine details are better recovered than the other two methods. Our color model performs very well with light skin colors while it lacks precision with darker skin tones where FitMe seems to perform better

Our approach captures subtle geometric details, including wrinkles and folds, in contrast to other methods, which struggle to reproduce these nuances. Notably, the shadows caused by wrinkles (see the first row) are baked in the albedo for both FitMe and Relightify, since they were not captured in the geometry, which cause unrealistic wrinkles in other light conditions. 
Our method is better at  capturing the facial structure (cheek, nose, eyes) and expression, resulting in a likeness that more faithfully resembles the original image. 
For darker skin tones, while our method has a better shape estimation, it tends to yield lighter skin tones.

\section{Limitations}
Our method does not explicitly model external occluders (e.g. eyeglass, hair) which get baked into the texture maps. Also, our method is more successful in preserving the skin tones for caucasian subjects. This is mainly due to the representation imbalance in our in-the-wild dataset. As shown in \cite{Maluke2022race}, FFHQ is composed of around 69\% white, 4\% black, and 27\% of other races. This biases the model towards generating lighter skin tones. Interestingly, we do not notice the same bias in the 3D shape reconstruction.
%The proposed system has difficulty dealing with glasses and other accessories like piercings. Since we do not explicitly model these objects, they get baked into the texture maps. We also get artifacts from skin zones occluded by hairs. 
%We also noticed that our method is more successful in preserving the skin tones for caucasian subjects. This is mainly due to the representation imbalance in our in-the-wild dataset. As shown in \cite{Maluke2022race}, FFHQ is composed of around 69\% white, 4\% black, and 27\% of other races. This biases the model towards generating lighter skin tones. Interestingly, we do not notice a bias in the 3D shape reconstruction.
%Again the hairs get baked in to the different maps resulting is weird skin patterns.

\section{Conclusion}
In this work, we presented MoSAR, a complete framework for creating realistic, relightable avatars from a single portrait image. It produces detailed geometry and skin reflectance maps at 4K resolution, compatible with modern rendering engines. We proposed a semi-supervised training procedure that includes a novel differentiable shading formulation that allows for estimating ambient occlusion and translucency. Our experiments highlighted the benefit of our semi-supervised training and showed that we obtain competitive results for geometry estimation using non-linear morphable models. This amounts to realistic and visually appealing renderings when compared to existing state-of-the-art methods. Finally, we introduced a dataset containing intrinsic facial attributes for $10$k subjects of the FFHQ-UV dataset to accelerate research on this field. For future work, we intend to collect more balanced dataset to reduce bias in our results. We want to model external occluders and reduce artifacts caused by hairs.
%The proposed semi-supervised training allows for capturing detailed geometry and skin reflectance maps at 4K resolution, compatible with modern rendering engines. 
%Our novel differentiable formulation allows for estimating additional maps (Ambient occlusion and Translucency).  % Improve
%It also allow for disentanglement of the diffuse and ambient occlusion shading granting artists greater control over these aspects. 

\section{Acknowledgments}
We would like to thank Dr. Alexandros Lattas and Foivos Papantoniou for running the FitMe and Relightify methods on the selected subjects. We thank the authors of DECA, HRN and Deep3D for sharing their code source publicly. We thank Amaury Depierre for assistance with the super-resolution network.

%This enables a significant improvement in the quality of facial reconstructions from in-the-wild images.
%Compared to existing methods, our approach produces more realistic and visually appealing renderings, reconstructing a more detailed geometry and reflectance.
%Finally, we  proposed a new dataset that contains intrinsic facial attributes for $10 000$ subjects of the FFHQ-UV dataset. 
%This is the first dataset to provide intrinsic facial attributes at scale, enabling new research directions on this field.

%\ad{TODO: talk about the new dataset at least here. TODO: say that we are the first to estimate AO and thk. }
%We presented a novel method for generating realistic, relightable avatars from a single portrait image. By jointly training with light stage and in-the-wild data, the proposed method is able to generate high quality geometry and reflectance maps, from an input image captured in arbitrary conditions.  Our ablation studies show that the proposed differentiable shader plays a critical role in disentangling the intrinsic reflectance components of the skin, which is necessary for rendering the avatar in different light conditions. 

%\input{7.comparison}
%\input{8.conclusion}

{\small
\bibliographystyle{ieeenat_fullname}
\bibliography{ms}
}

\end{document}

\maketitle


\section{Hyperparameters}

We use the following hyperparameters for training the 3D Geometry Reconstruction networks: $lr = 0.00005$, $\lambda_\text{photo} = 0.1$, $\lambda_\text{landmark} = 0.001$, $\lambda_\text{lap} = 0.1$, $\lambda_\text{light} = 0.001$, $\lambda_\text{exp} =  0.1$, $\lambda_\text{alb} =  0.1$, $\lambda_\text{supervised} = 1.$, $\lambda_\text{nrm}  = 0.1$.

The texture completion and light normalization networks are both trained with $lr = 0.001$.

The displacement map estimation is trained with $lr = 0.0001$. The texture estimation networks are trained with: $lr = 0.0001$, $\lambda_\text{shading} = 0.5$, $\lambda_\text{sup} = 1$, $\lambda_\text{GAN} = 1$. 

All modules were implemented in Pytorch, and trained on two CUDA-enabled GPUs with 24 GB RAM. Used the Adam optimizer \cite{kingma2014adam} for training all networks.

\section{FFHQ-UV-Intrinsics}
In this section, we describe the process of generating our new dataset named \emph{FFHQ-UV-Intrinsics}, built from the publicly available dataset FFHQ-UV \cite{Bai_2023_CVPR}. 
The FFHQ-UV dataset is composed of texture maps of $1K$ resolution, for subjects sampled from the latent space of StyleGAN. These texture contains evenly illuminated face images. However, light, geometry and skin reflectance information are entangled in the same texture making them less suitable for relighting.

To obtain the intrinsic face attributes, we first re-targeted the texture maps to our own topology and resize them to $512 \times 512$. Next, we apply the proposed \emph{light normalization} and \emph{Intrinsic texture maps estimation} steps. We then upscale these texture maps to $1$K resolution and retarget them back to their original topology. 
%We note that we do not release the $4K$ version of these texture due to the online storage limit.

%Figure \ref{fig:ffhq} shows three samples of the original FFHQ-UV dataset, and the resulting texture maps obtained by our decomposition process in our topology. 

The resulting dataset, \emph{FFHQ-UV-Intrinsics}, is being publicly released for the research community. The dataset contains diffuse, specular, ambient occlusion, translucency and normal maps for $10$K subjects. This is the first dataset that offer rich intrinsic face attributes at high resolution and at large scale, with the aim of advancing research in this field. 
%  \begin{figure*}
%     \centering
%     \includegraphics[width=0.8\textwidth,height=\textheight,keepaspectratio]{suppl_figures/ffhq-intrinsics.pdf}
%     \caption{Samples from the proposed \emph{FFHQ-UV-Intrinsics} dataset. For each subject, we estimate the light normalized  $\mathcal{M}$, diffuse $\mathcal{D}$, specular $\mathcal{S}$, normal $\mathcal{N}$, ambient occlusion $\mathcal{A}$ and translucency $\mathcal{T}$ texture maps.}
%     \label{fig:ffhq}
% \end{figure*}

\section{Comparisons on additional subjects}
\begin{figure*}
    \centering
    \includegraphics[width=\textwidth,height=\textheight,keepaspectratio]{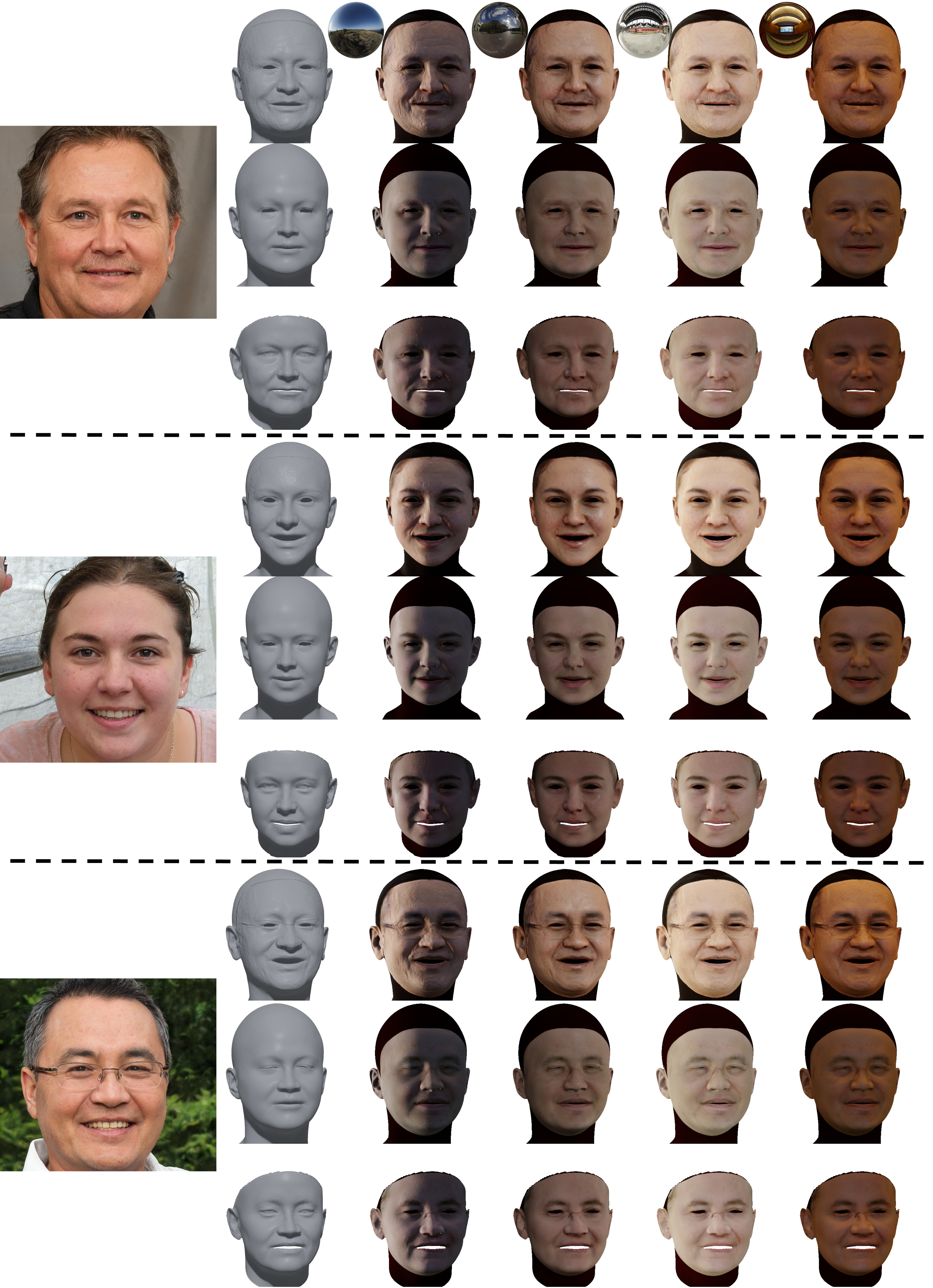}
    \caption{Comparison of the estimated geometry and renders, under 4 lighting conditions, between our method (first row),  FitMe \cite{lattas2023fitme} (second row) and Relightify \cite{papantoniou2023relightify} (third row)}
    \label{fig:relight1}
\end{figure*}

\begin{figure*}
    \centering
    \includegraphics[width=\textwidth,height=\textheight,keepaspectratio]{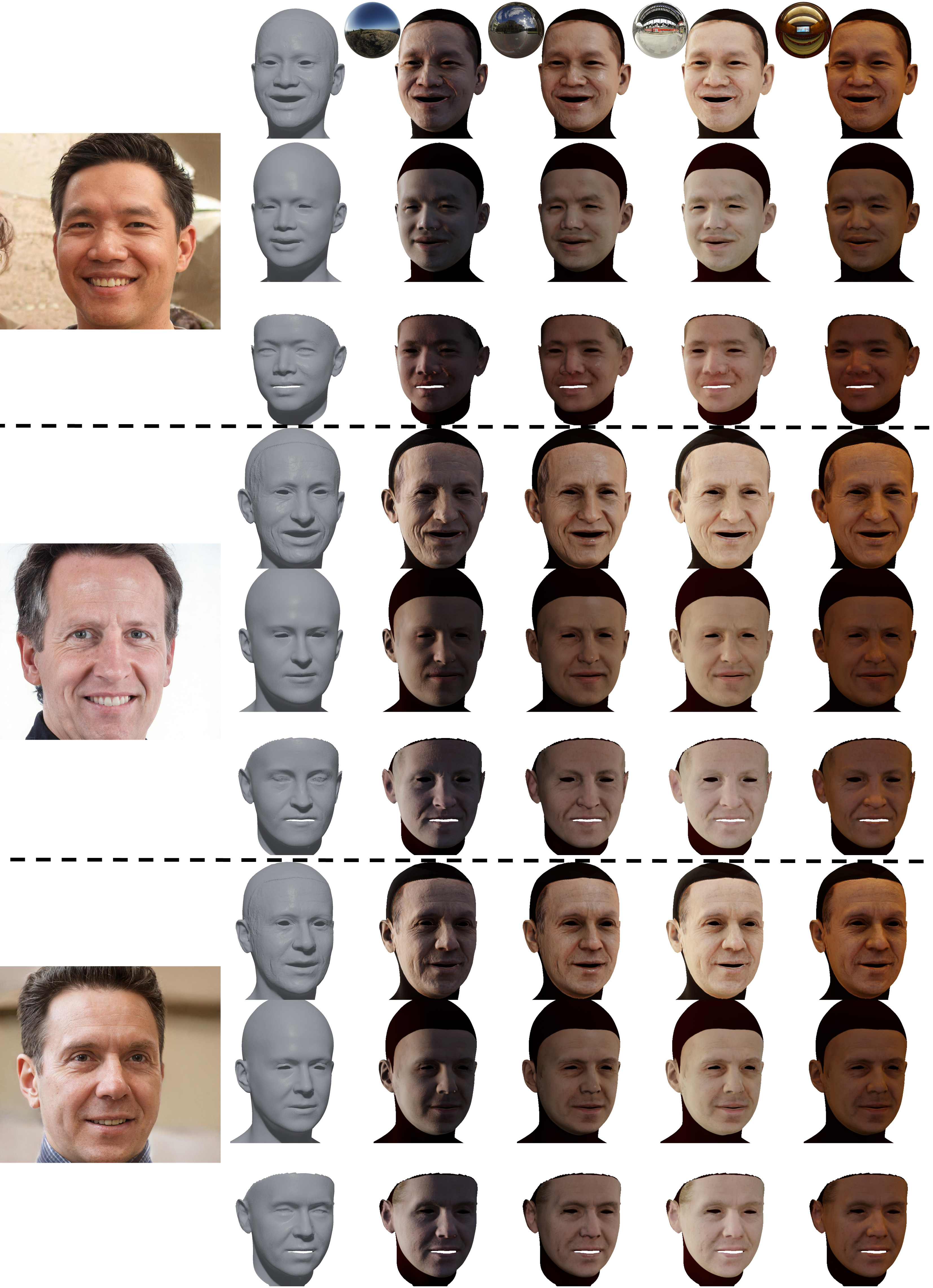}
    \caption{Comparison of the estimated geometry and renders, under 4 lighting conditions, between our method (first row),  FitMe \cite{lattas2023fitme} (second row) and Relightify \cite{papantoniou2023relightify} (third row) }
    \label{fig:relight2}
\end{figure*}

\begin{figure*}
    \centering
    \includegraphics[width=\textwidth,height=\textheight,keepaspectratio]{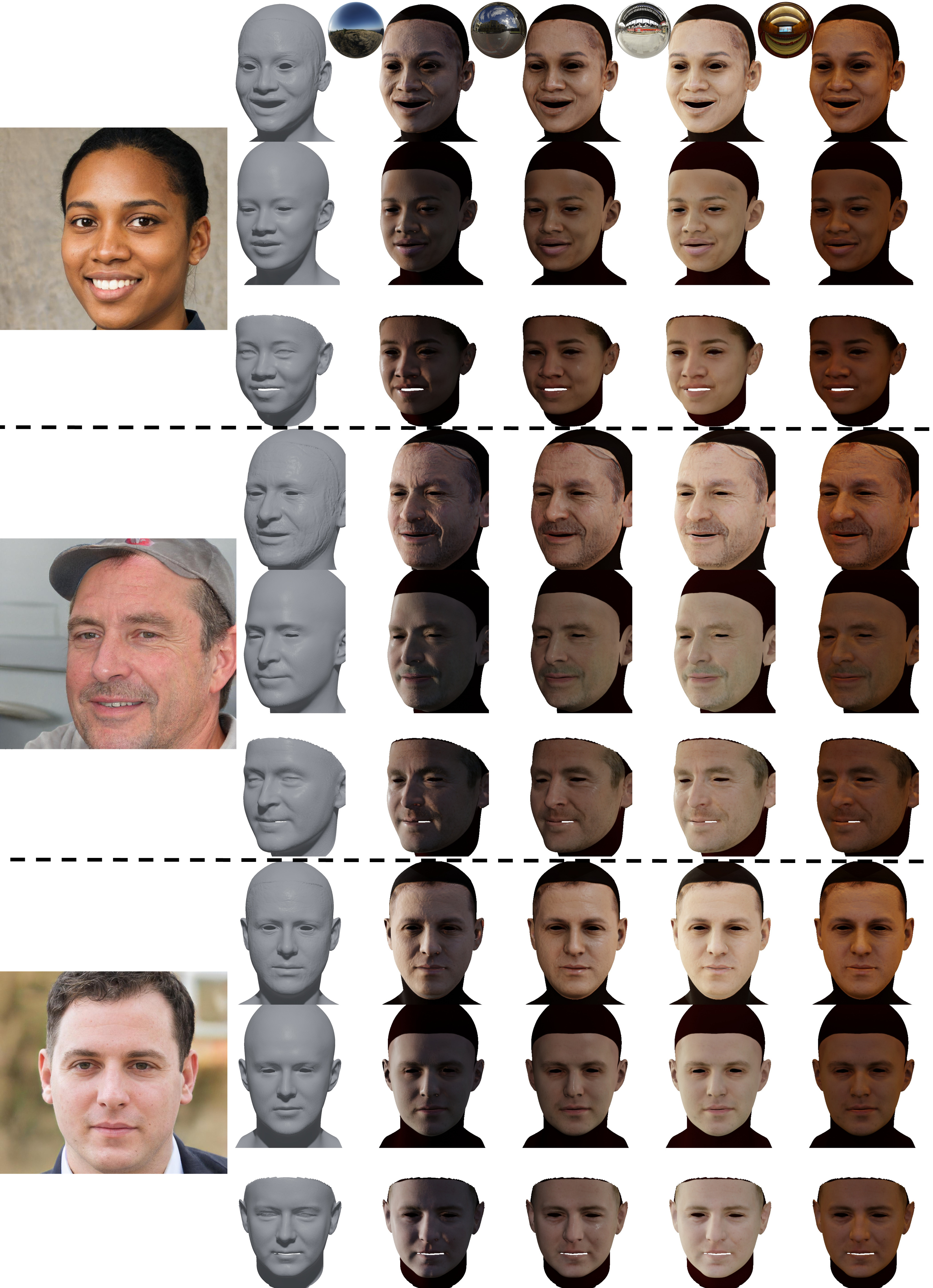}
    \caption{ Comparison of the estimated geometry and renders, under 4 lighting conditions, between our method (first row),  FitMe \cite{lattas2023fitme} (second row) and Relightify \cite{papantoniou2023relightify} (third row)}
    \label{fig:relight3}
\end{figure*}

\begin{figure*}
    \centering
    \includegraphics[width=\textwidth,height=\textheight,keepaspectratio]{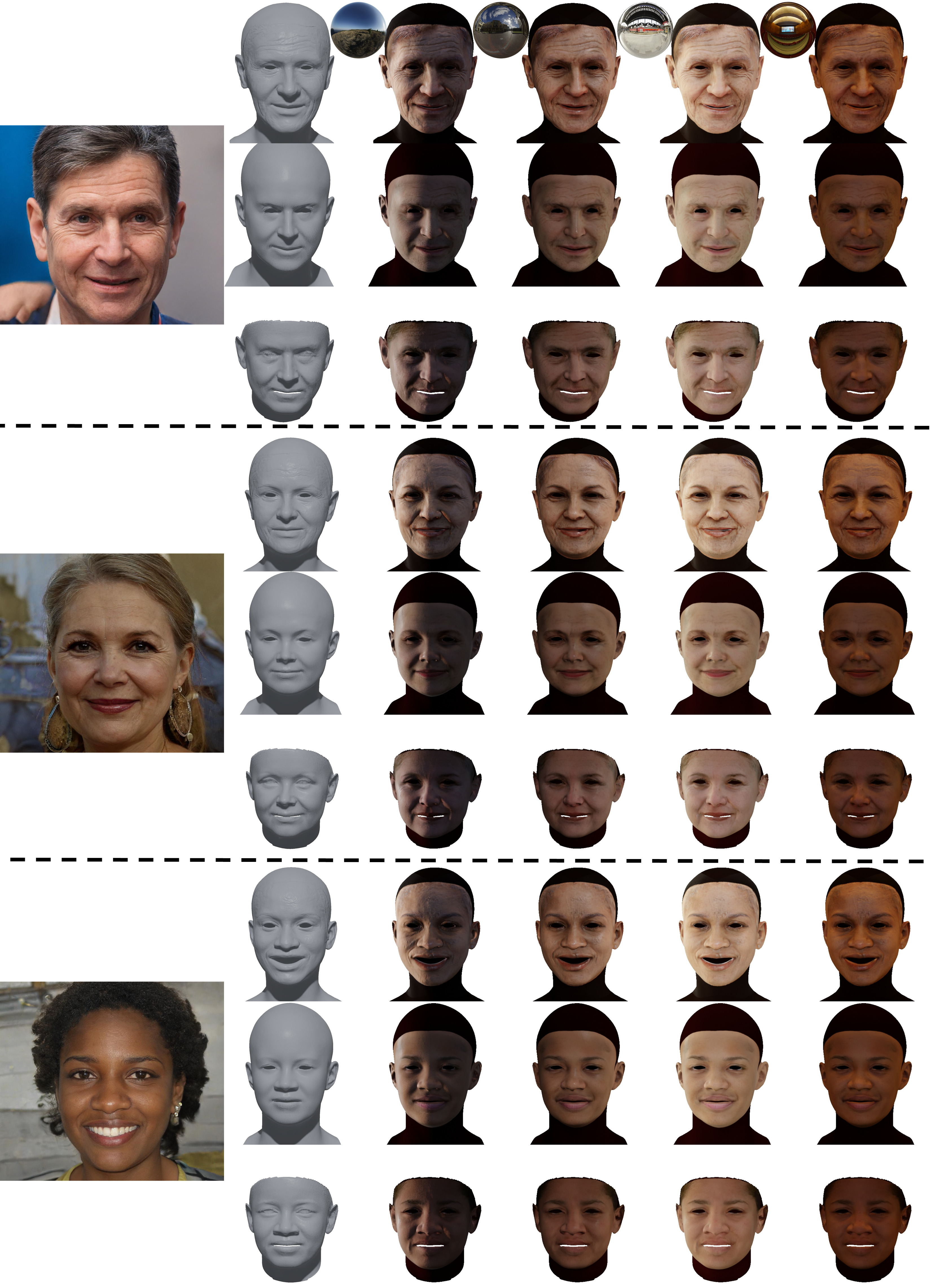}
    \caption{ Comparison of the estimated geometry and renders, under 4 lighting conditions, between our method (first row),  FitMe \cite{lattas2023fitme} (second row) and Relightify \cite{papantoniou2023relightify} (third row)}
    \label{fig:relight4}
\end{figure*}

\begin{figure*}
    \centering
    \includegraphics[width=\textwidth,height=\textheight,keepaspectratio]{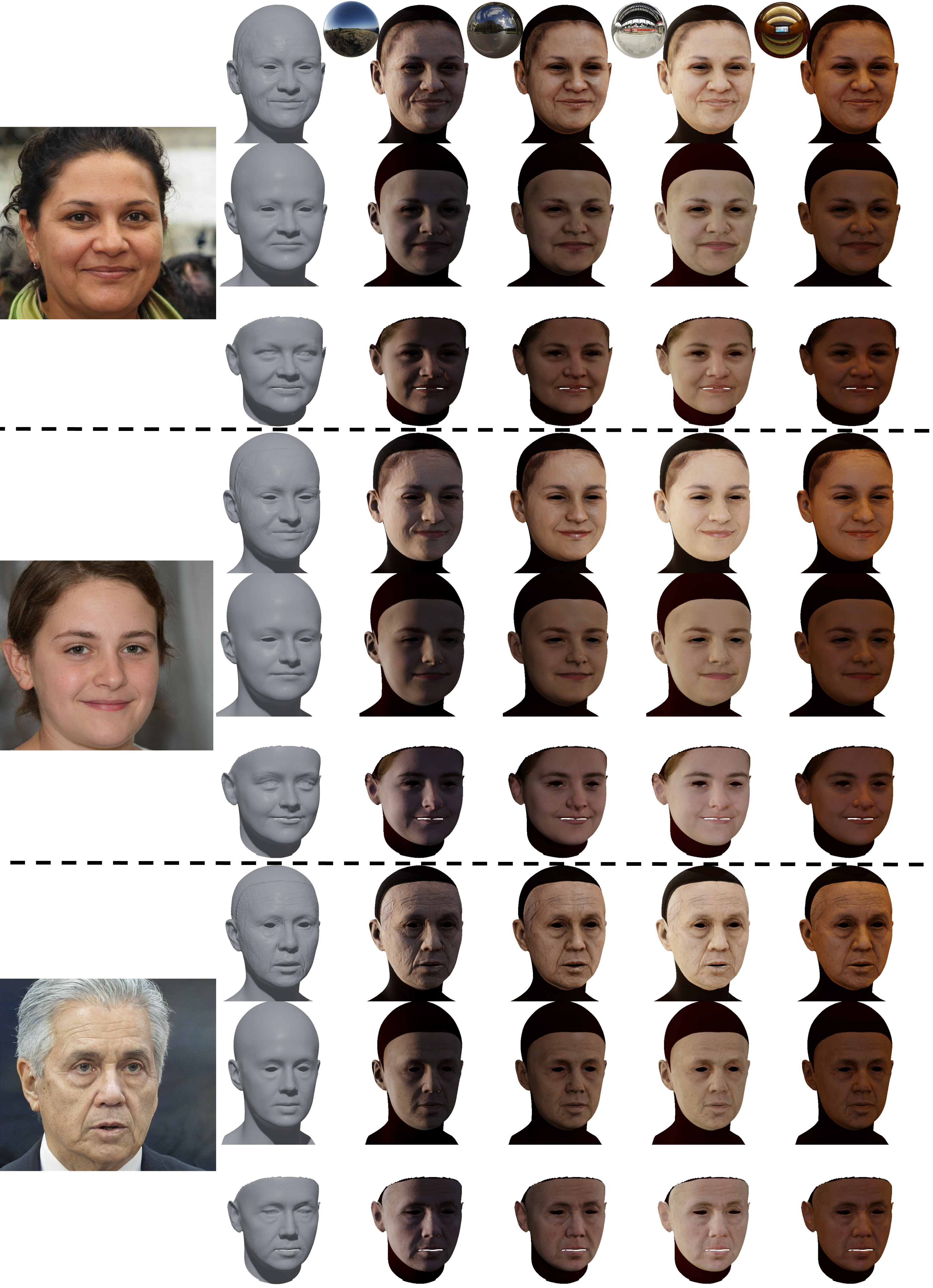}
    \caption{ Comparison of the estimated geometry and renders, under 4 lighting conditions, between our method (first row),  FitMe \cite{lattas2023fitme} (second row) and Relightify \cite{papantoniou2023relightify} (third row)}
    \label{fig:relight5}
\end{figure*}

\begin{figure*}
    \centering
    \includegraphics[width=\textwidth,height=\textheight,keepaspectratio]{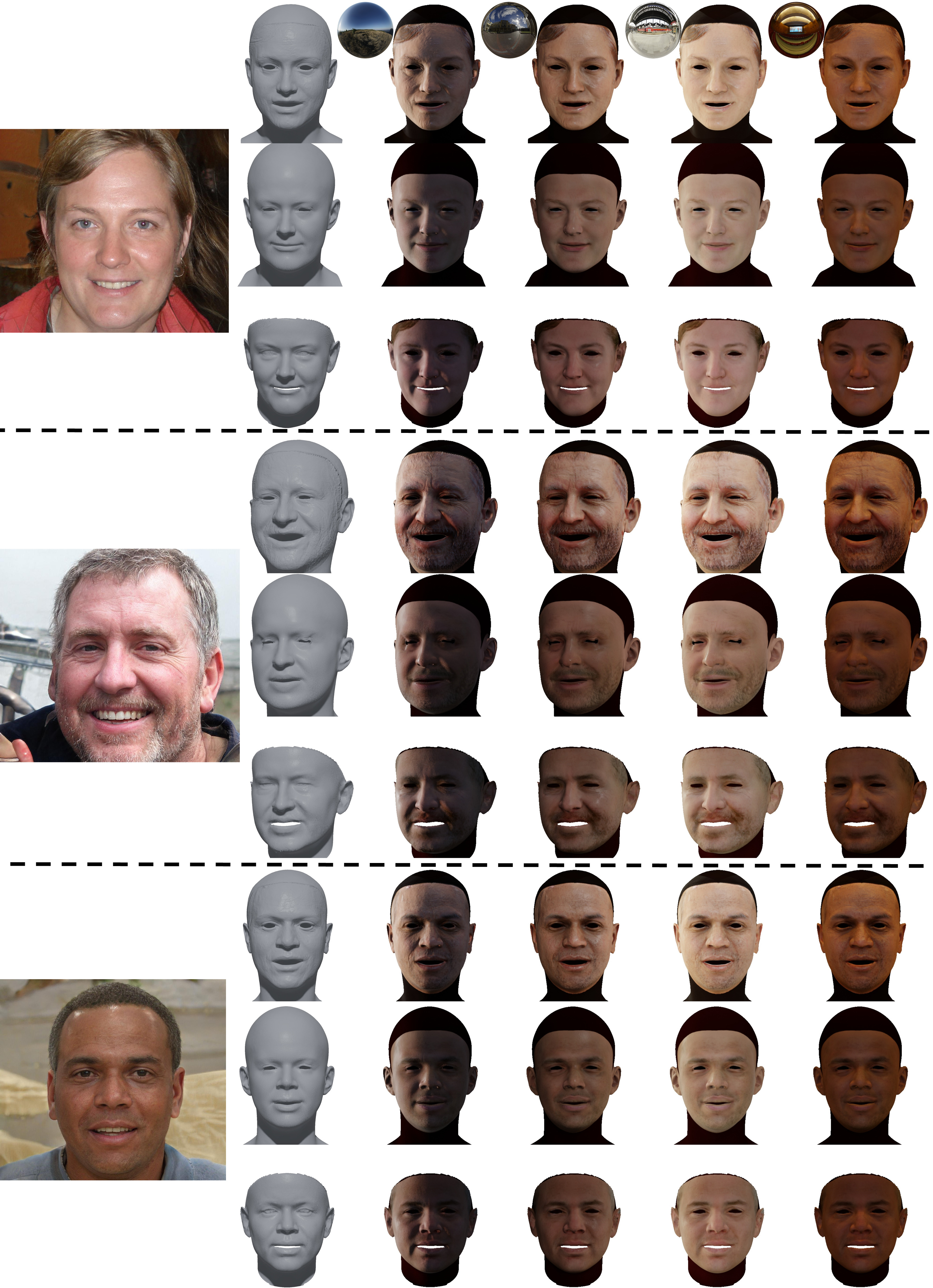}
    \caption{Comparison of the estimated geometry and renders, under 4 lighting conditions, between our method (first row),  FitMe \cite{lattas2023fitme} (second row) and Relightify \cite{papantoniou2023relightify} (third row) }
    \label{fig:relight6}
\end{figure*}

\begin{figure*}
    \centering
    \includegraphics[width=\textwidth,height=\textheight,keepaspectratio]{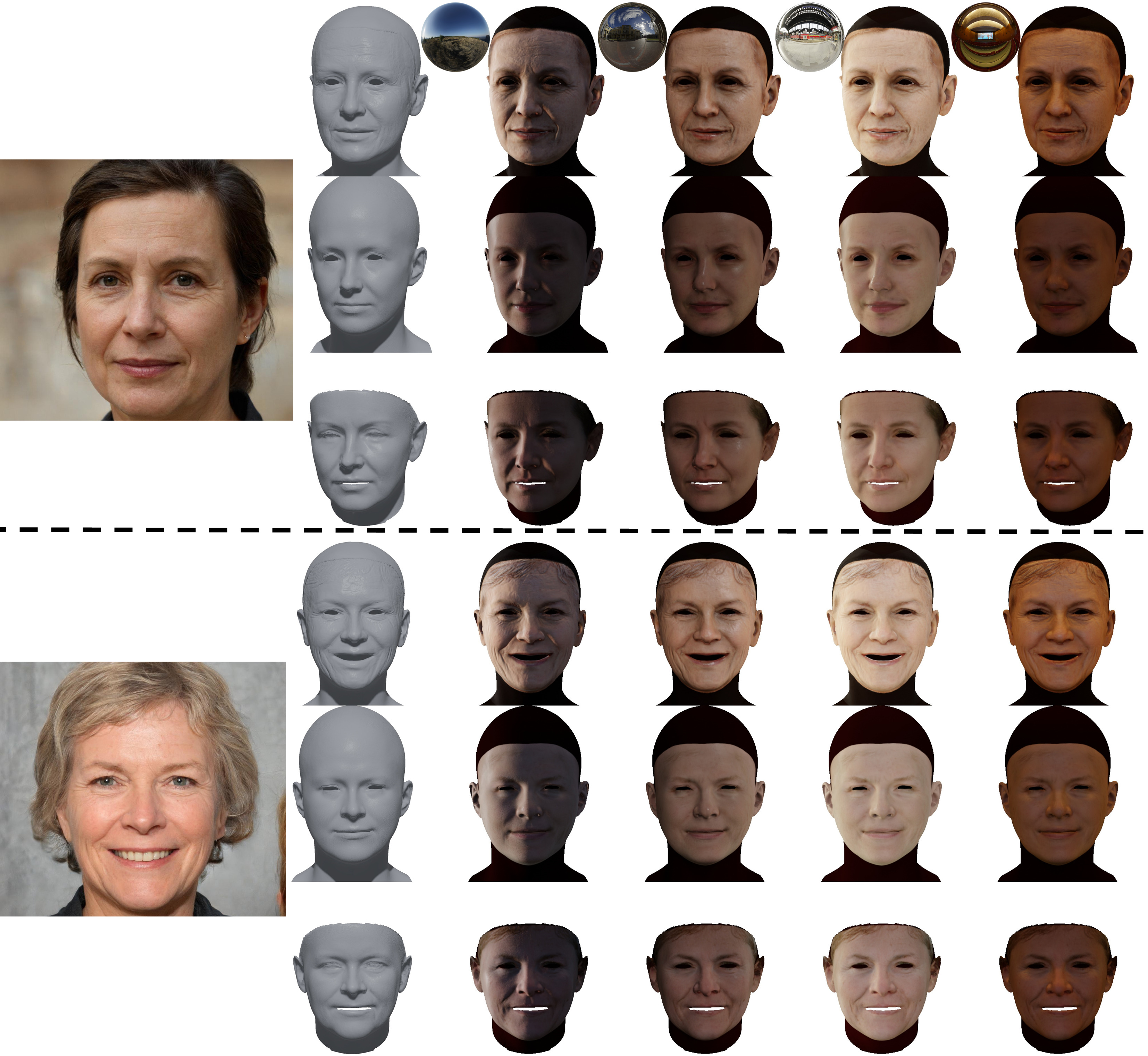}
    \caption{Comparison of the estimated geometry and renders, under 4 lighting conditions, between our method (first row),  FitMe \cite{lattas2023fitme} (second row) and Relightify \cite{papantoniou2023relightify} (third row) }
    \label{fig:relight7}
\end{figure*}

\Cref{fig:relight1,fig:relight2,fig:relight3,fig:relight4,fig:relight5,fig:relight6,fig:relight7} show additional comparisons between our method,  FitMe \cite{lattas2023fitme} (second row) and Relightify \cite{papantoniou2023relightify} (third row).  For every method, we show the estimated geometry and the rendering under 4 different environment maps. 

\begin{figure*}
    \centering
    \includegraphics[width=\textwidth,height=\textheight,keepaspectratio]{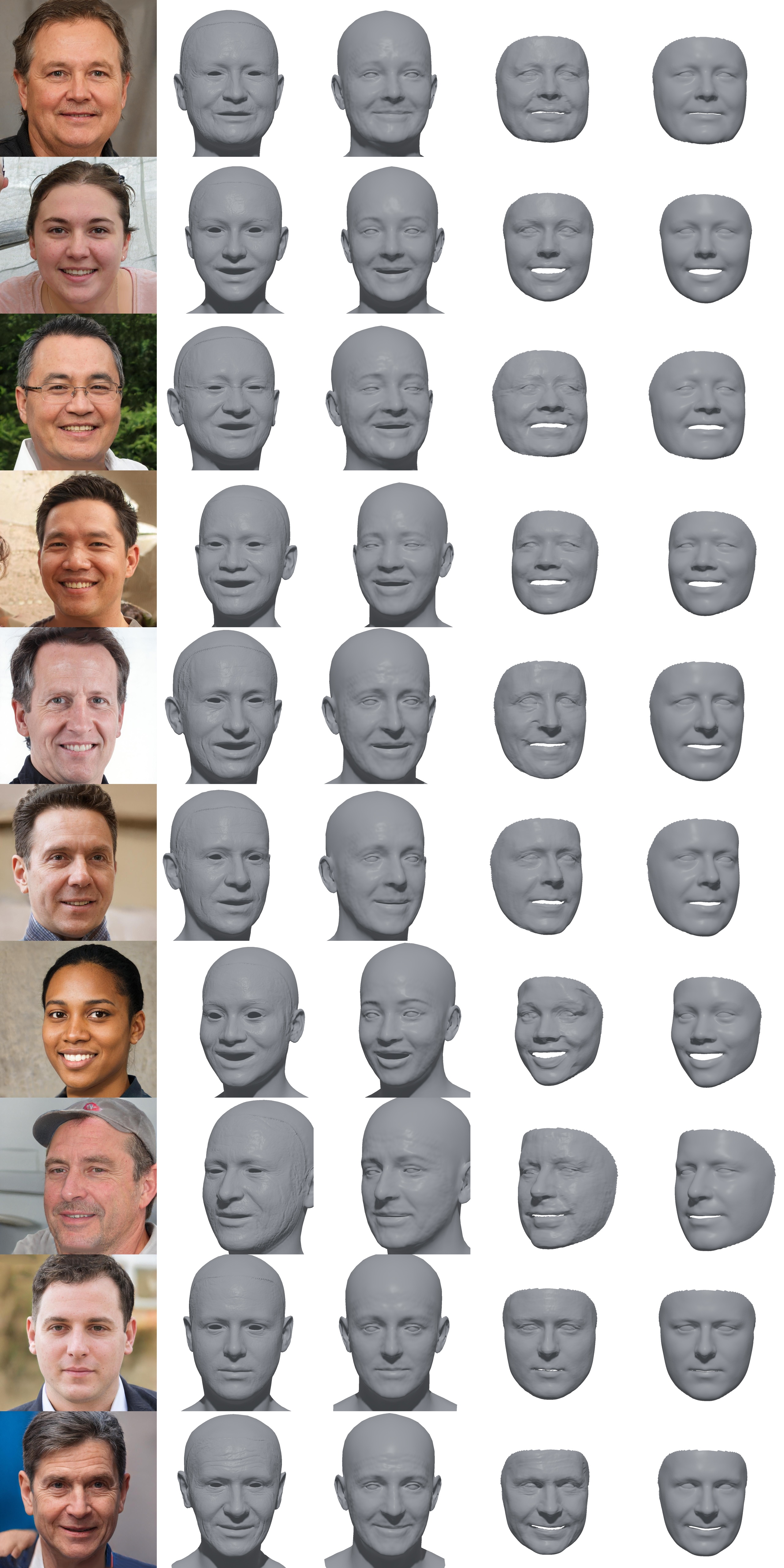}
    \caption{ Comparison of the estimated geometry between our method (second row), DECA \cite{feng2021learning} (third row), HRN \cite{lei2023hierarchical} (fourth row) and Deep3D \cite{deng2019accurate} (last row)}
    \label{fig:sota1}
\end{figure*}

\begin{figure*}
    \centering
    \includegraphics[width=\textwidth,height=\textheight,keepaspectratio]{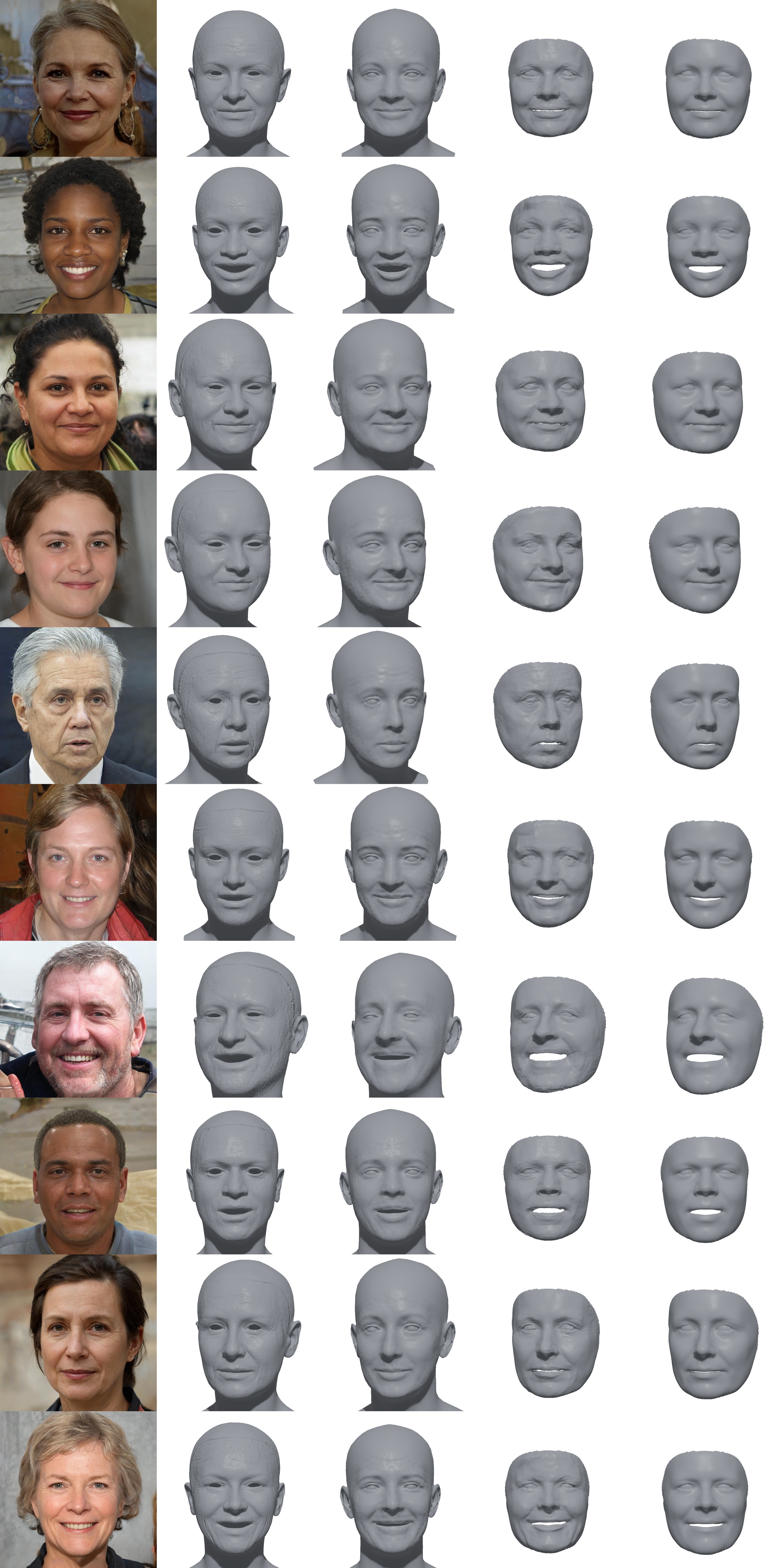}
    \caption{ Comparison of the estimated geometry between our method (second row), DECA \cite{feng2021learning} (third row), HRN \cite{lei2023hierarchical} (fourth row) and Deep3D \cite{deng2019accurate} (last row)}
    \label{fig:sota2}
\end{figure*}

Additionally, we perform geometry comparison on the same 20 subjects for methods that only estimates geometry. \Cref{fig:sota1,fig:sota2} show comparison of our estimated geometry against DECA \cite{feng2021learning} , HRN \cite{lei2023hierarchical} and Deep3D \cite{deng2019accurate}.



%\section{Intrinsic estimation comparison on light-stage data}

% \begin{figure*}
%     \centering
%     \includegraphics[width=\textwidth,height=\textheight,keepaspectratio]{suppl_figures/relight.pdf}
%     \caption{ }
%     \label{fig:relight}
% \end{figure*}
% \begin{figure*}
%     \centering
%     \includegraphics[width=\textwidth,height=\textheight,keepaspectratio]{suppl_figures/geom_first10_1k.jpg}
%     \caption{ }
%     \label{fig:geom1}
% \end{figure*}
% \begin{figure*}
%     \centering
%     \includegraphics[width=\textwidth,height=\textheight,keepaspectratio]{suppl_figures/geom_last10_1k.jpg}
%     \caption{ }
%     \label{fig:geom2}
% \end{figure*}
% \begin{figure*}
%     \centering
%     \includegraphics[width=\textwidth,height=\textheight,keepaspectratio]{suppl_figures/vs_sota_first10_1k.jpg}
%     \caption{ }
%     \label{fig:sota1}
% \end{figure*}
% \begin{figure*}
%     \centering
%     \includegraphics[width=\textwidth,height=\textheight,keepaspectratio]{suppl_figures/vs_sota_last10_1k.jpg}
%     \caption{ }
%     \label{fig:sota2}
% \end{figure*}
\clearpage
\clearpage
{
    \small
    \bibliographystyle{ieeenat_fullname}
    \bibliography{supplement}
}